\documentclass[10pt,twocolumn,letterpaper]{article}

\usepackage{cvpr}
\usepackage{times}
\usepackage{epsfig}
\usepackage{graphicx}
\usepackage{amsmath}
\usepackage{amssymb}
\usepackage[percent]{overpic}
\usepackage{times}
\usepackage{epsfig}
\usepackage{graphicx}
\usepackage{float}
\usepackage{wrapfig}
\usepackage{amsmath,amssymb,amsthm}
\usepackage{algorithm,algorithmicx,algpseudocode}
\usepackage{bm,xspace}
\usepackage{comment}
\usepackage{verbatim}
\usepackage{multirow}
\usepackage{balance}
\usepackage{url}
\usepackage{booktabs}
\usepackage{etoolbox,siunitx}
\usepackage{calc}
\usepackage{pifont,hologo}
\usepackage{dblfloatfix}
\usepackage{pdfpages}
\usepackage{multido}

\usepackage[table,xcdraw,dvipsnames]{xcolor}
\definecolor{LightCyan}{rgb}{0.88,1,1}
\definecolor{LightYellow}{rgb}{1,1,0.7}
\definecolor{LightGreen}{rgb}{0.4, 1, 0.6}

\def\netname{$\Omega$Net}

\usepackage[pagebackref=true,breaklinks=true,letterpaper=true,colorlinks,bookmarks=false]{hyperref}

\cvprfinalcopy %

\def\cvprPaperID{7782}

\begin{document}

\title{Distilled Semantics for Comprehensive Scene Understanding from Videos}

\author{Fabio Tosi\thanks{Joint first authorship.} \hspace*{1cm} Filippo Aleotti$^*$ \hspace*{1cm} Pierluigi Zama Ramirez$^*$ \\
Matteo Poggi \hspace*{1cm} Samuele Salti \hspace*{1cm} Luigi Di Stefano \hspace*{1cm} Stefano Mattoccia \\
Department of Computer Science and Engineering (DISI)\\
University of Bologna, Italy\\
{\tt\small $^*$\{fabio.tosi5, filippo.aleotti2, pierluigi.zama\}@unibo.it}
}

\makeatletter
\g@addto@macro\@maketitle{
  \begin{figure}[H]
  \setlength{\linewidth}{\textwidth}
  \setlength{\hsize}{\textwidth}
  \vspace{-7mm}
  \centering
  \renewcommand{\tabcolsep}{1pt} 
 	\begin{tabular}{ccc}
 	\begin{overpic}[width=0.25\textwidth]{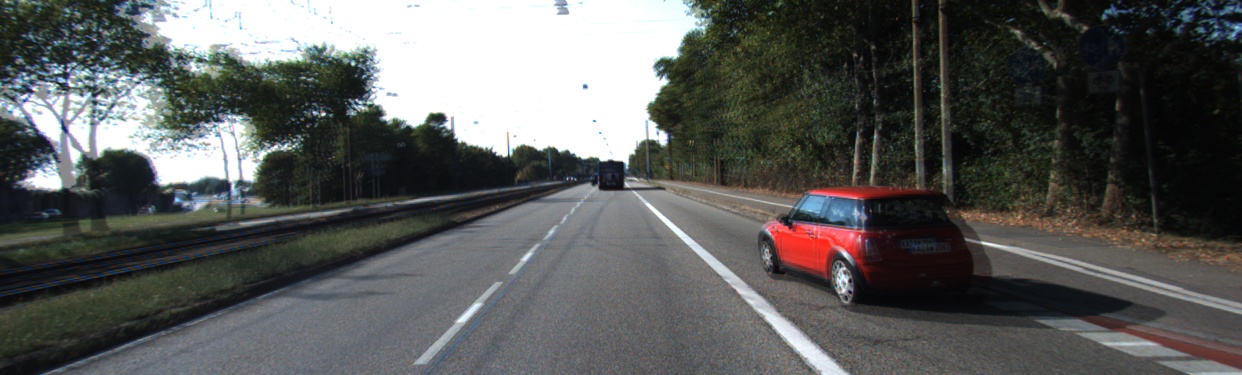}
    \put (0,3) {\colorbox{white}{$\displaystyle\textcolor{black}{\text{(a)}}$}}
    \end{overpic} 
    \begin{overpic}[width=0.25\textwidth]{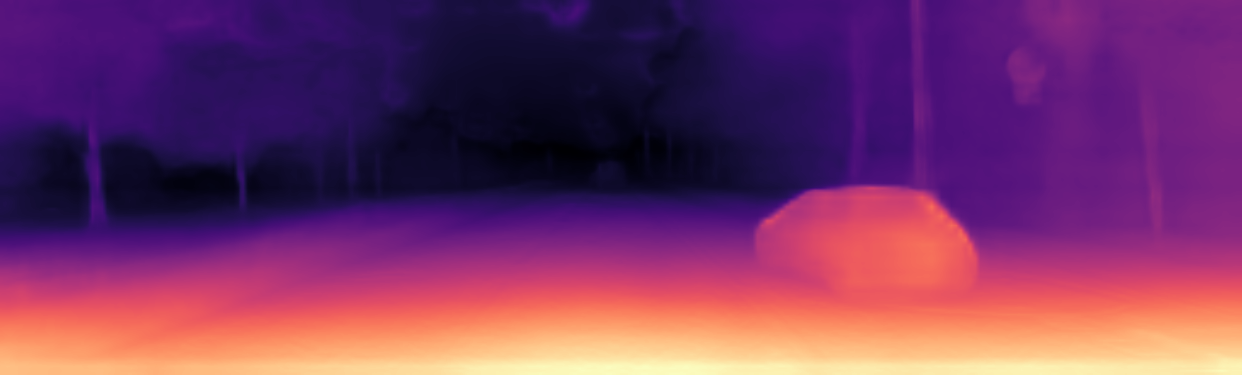}
    \put (0,3) {\colorbox{white}{$\displaystyle\textcolor{black}{\text{(b)}}$}}
    \end{overpic}
    \begin{overpic}[width=0.25\textwidth]{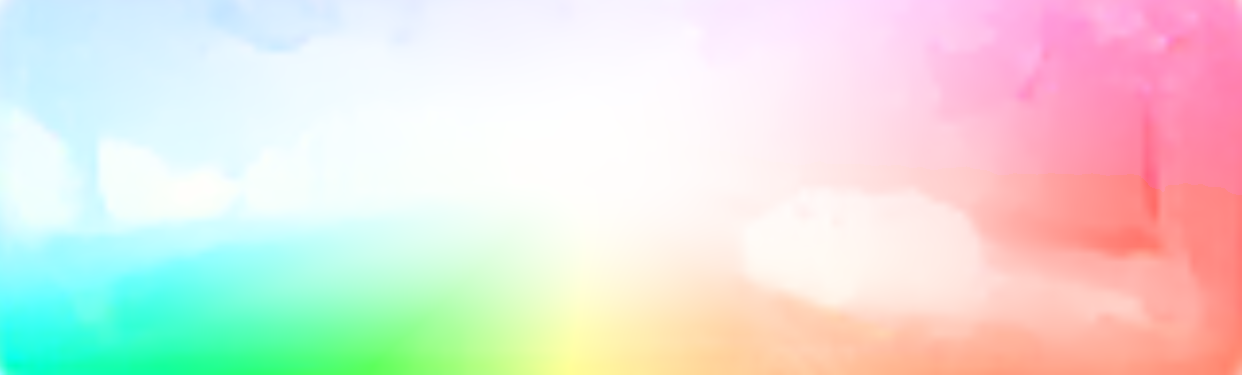}
    \put (0,3) {\colorbox{white}{$\displaystyle\textcolor{black}{\text{(c)}}$}}
    \end{overpic} \\

 	\begin{overpic}[width=0.25\textwidth]{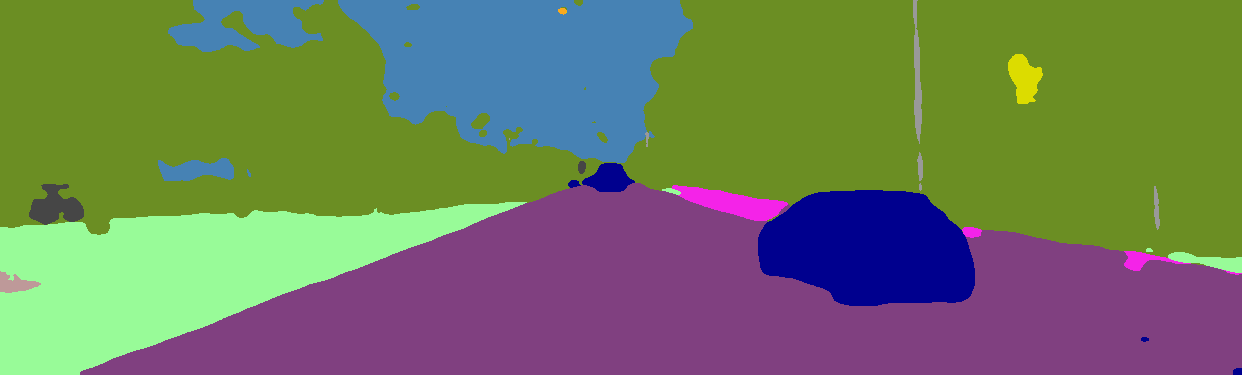}
    \put (0,3) {\colorbox{white}{$\displaystyle\textcolor{black}{\text{(d)}}$}}
    \end{overpic} 
    \begin{overpic}[width=0.25\textwidth]{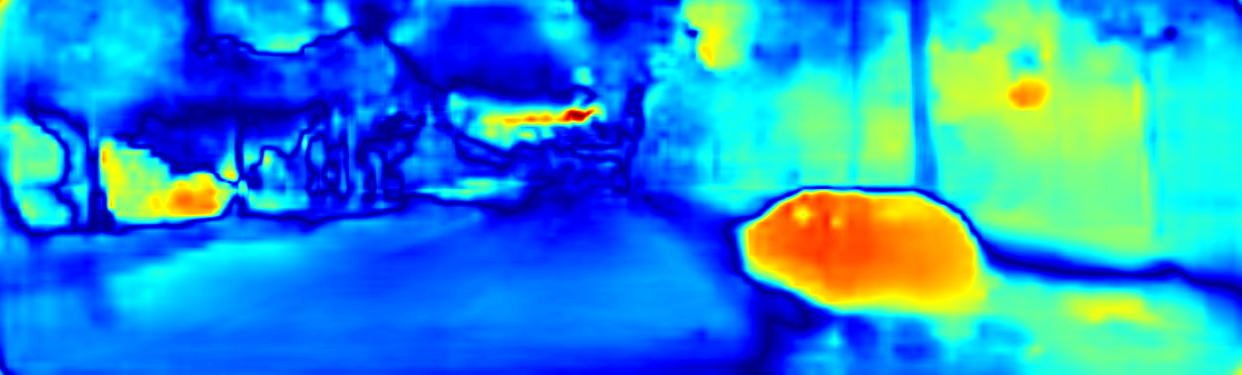}
    \put (0,3) {\colorbox{white}{$\displaystyle\textcolor{black}{\text{(e)}}$}}
    \end{overpic}
    \begin{overpic}[width=0.25\textwidth]{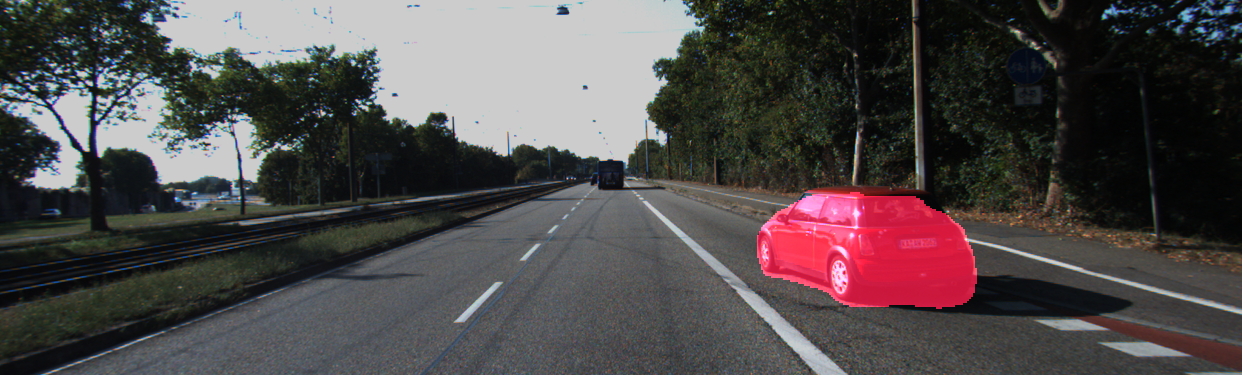}
    \put (0,3) {\colorbox{white}{$\displaystyle\textcolor{black}{\text{(f)}}$}}
    \end{overpic} \\   
    \end{tabular}
    \vspace{1mm}
    \caption{Given an input monocular video (a),  our network can provide the following outputs in real-time: depth (b), optical flow (c), semantic labels (d), per-pixel motion probabilities (e), motion mask (f).}
  \label{fig:abstract}
  \end{figure}
}
\makeatother

\maketitle

\begin{abstract}
Whole understanding of the surroundings is paramount to autonomous systems. Recent works have shown that deep neural networks can learn geometry (depth) and motion (optical flow) from a monocular video without any explicit supervision from ground truth annotations, particularly hard to source for these two tasks. In this paper, we take an additional step toward holistic scene understanding with monocular cameras by learning depth and motion alongside with semantics, with supervision for the latter provided by a pre-trained network distilling proxy ground truth images. 
We address the three tasks jointly by a) a novel training protocol based on knowledge distillation and self-supervision and b) a compact network architecture which enables efficient scene understanding on 
both power hungry GPUs and low-power embedded platforms.
We thoroughly assess the performance of our framework and show that it yields state-of-the-art results for monocular depth estimation, optical flow and motion segmentation. 

\end{abstract}


\section{Introduction}
\label{sec:introduction}

What information would an autonomous agent be keen to gather from its sensory sub-system to tackle tasks like navigation and interaction with the explored environment? It would need to be informed about the geometry of the surroundings and the type of objects therein, and likely better know which of the latter are actually moving and how they do so. What if all such cues may be provided by as simple a sensor as a single RGB camera? 

Nowadays, deep learning is advancing the state-of-the-art in classical computer vision problems at such a quick pace that single-view holistic scene understanding seems to be no longer out-of-reach. 
Indeed, highly challenging problems such as monocular depth estimation and optical flow can nowadays be addressed successfully by deep neural networks, often through unified architectures \cite{Yin_CVPR_2018,Ranjan_CVPR_2019,Zou_ECCV_2018}. Self-supervised learning techniques have yielded further major achievements  \cite{Zhou_2017_CVPR,Meister_AAAI_2018} by enabling effective training of deep networks without annotated images. In fact, labels are hard to source for depth estimation due to the need of active sensors and  manual filtering, and are even more cumbersome in the case of optical flow.
Concurrently, semi-supervised approaches \cite{Ramirez_ACCV_2018,Chen_CVPR_2019} proved how a few semantically labelled images can improve monocular depth estimation significantly. These works have also highlighted how, while producing per-pixel class labels is tedious yet feasible for a human annotator, manually endowing  images with depth and optical flow ground-truths is prohibitive.

In this paper, we propose the first-ever framework for comprehensive scene understanding from monocular videos. As highlighted in Figure \ref{fig:abstract}, our multi-stage network architecture, named \netname{}, can predict depth, semantics, optical flow, per-pixel motion probabilities and motion masks. This comes alongside with estimating the pose between adjacent frames for an uncalibrated camera, whose intrinsic parameters are also estimated. Our training methodology leverages on self-supervision,  knowledge distillation and multi-task learning. In particular, peculiar to our proposal and key to performance is distillation of proxy semantic labels gathered from a state-of-the-art pre-trained model \cite{autodeeplab_CVPR_2019}  within a self-supervised and multi-task learning procedure addressing depth, optical flow and motion segmentation. Our training procedure also features a novel and effective self-distillation schedule for optical flow mostly aimed at handling occlusions and relying on tight integration of rigid flow, motion probabilities and semantics.
Moreover, \netname{} is lightweight, counting less than $8.5$M parameters, and fast, as it can run at nearly 60 FPS and 5 FPS on an NVIDIA Titan Xp and a Jetson TX2, respectively. 
As vouched by thorough experiments, the main contributions of our work can be summarized as follows:

$\bullet$ The first real-time network for joint prediction of depth, optical flow, semantics and motion segmentation from monocular videos

$\bullet$ A novel training protocol relying on proxy semantics and self-distillation to effectively address the self-supervised multi-task learning problem 

$\bullet$ State-of-the-art self-supervised monocular depth estimation, largely improving accuracy at long distances

$\bullet$ State-of-the-art optical flow estimation among monocular multi-task frameworks, thanks to our novel occlusion-aware and semantically guided training paradigm

$\bullet$ State-of-the-art motion segmentation by joint reasoning about optical-flow and semantics


\section{Related Work}
\label{sec:related}
We review previous works relevant to our proposal. 

\textbf{Monocular depth estimation.} At first, depth estimation was tackled as a supervised \cite{Eigen_2014, Laina_3DV_2016} or semi-supervised task \cite{Kuznietsov_CVPR_2017}. 
Nonetheless, self-supervision from image reconstruction is now becoming the preferred paradigm to avoid hard to source labels. Stereo pairs \cite{Garg_ECCV_2016, Godard_CVPR_2017} can provide such supervision and enable scale recovery, with further improvements achievable by leveraging on  trinocular assumptions \cite{Poggi_3DV_2018}, proxy labels from SGM \cite{Tosi_CVPR_2019, Watson_ICCV_2019} or guidance from visual odometry \cite{vomonodepth19}.
Monocular videos \cite{Zhou_2017_CVPR} are a more flexible alternative, although they do not allow for scale recovery and mandate learning camera pose alongside with depth. Recent developments of this paradigm deal with  differentiable direct visual odometry \cite{Wang_CVPR_2018} or ICP  \cite{Mahjourian_CVPR_2018} and  normal consistency \cite{Yang_CVPR_2018}.
Similarly to our work, \cite{Yin_CVPR_2018,Zou_ECCV_2018,Chen_ICCV_2019, Ranjan_CVPR_2019, Yang_ECCV_Workshops_2018, Luo_EPC++_2018} model rigid and non-rigid components using the projected depth, relative camera transformations, and optical flow to handle independent motions, which can also be estimated independently in the 3D space \cite{Casser_AAAI_2019, Xu_IJCAI_2019}. In \cite{Gordon_ICCV_2019}, the authors show how to learn  camera intrinsics together with depth and egomotion to enable training on any unconstrained video. In \cite{Godard_ICCV_2019, zhou2019unsupervised, Bian_NeurIPS_2019}, reasoned design choices such as a minimum reprojection loss between frames, self-assembled attention modules and auto-mask strategies to handle static camera or dynamic objects proved to be very effective.
Supervision from stereo and video have also been combined \cite{Zhan_CVPR_2018,Godard_ICCV_2019}, possibly improved by means of proxy supervision from stereo direct sparse odometry \cite{Yang_ECCV_2018}. Uncertainty modeling for self-supervised monocular depth estimation has been studied in \cite{Poggi_2020_CVPR}.
Finally, lightweight networks aimed at real-time performance on low-power systems  have been proposed within  self-supervised \cite{Poggi_IROS_2018,DATE_2019} as well as supervised \cite{Diana_ICRA_2019} learning paradigms. 
 
\textbf{Semantic segmentation.}
Nowadays, fully convolutional neural networks \cite{Long_CVPR_2015} are the standard approach for semantic segmentation. Within this framework,  multi-scale context modules and proper architectural choices are crucial to performance. The former rely on spatial pyramid pooling \cite{He_PAMI_2015,Zhao_CVPR_2017} and atrous convolutions \cite{Chen_2017, Chen_PAMI_2017, Chen_ECCV_2018}. As for the latter, popular backbones \cite{Krizhevsky_NIPS_2012,Simonyan_ICLR_15,He_CVPR_2016} have been improved by more recent designs \cite{Huang_CVPR_2017,Chollet_CVPR_2017}.
While for years the encoder-decoder architecture has been the most popular choice \cite{Ronneberger_2015, Badrinarayanan_PAMI_2017}, recent trends in Auto Machine Learning (AutoML) \cite{autodeeplab_CVPR_2019, Chen_NIPS_2018} leverage on architectural search to achieve state-of-the-art accuracy. However, these latter have huge computational requirements.
An alternative research path deals with  real-time semantic segmentation networks. In this space, \cite{Paszke_2016} deploys a compact and efficient network architecture, \cite{Yu_ECCV_2018} proposes  a two paths network to attain fast inferences while capturing high resolution details. DABNet \cite{Li_BMVC_2019} finds an effective combinations of depth-wise separable filters and atrous-convolutions to reach a good trade-off between efficiency and accuracy. \cite{Li_CVPR_2019}  employs cascaded sub-stages to refine results while FCHardNet \cite{Chao_ICCV_2019} leverages on a new harmonic densely connected pattern to maximize the inference performance of larger networks.

\textbf{Optical flow estimation.}
The optical flow problem concerns estimation of the apparent displacement of pixels in consecutive frames, and it is useful in various applications such as, \eg, video editing \cite{Chang_CVPRW_2019, Jiang_CVPR_2018} and object tracking \cite{xiang2015learning}. Initially introduced by Horn and Schunck \cite{horn1981determining}, this problem has traditionally been tackled by variational approaches \cite{brox2004high,black1996robust, Revaud_CVPR_2015}. More recently, Dosovitskiy \etal \cite{Dosovitskiy_ICCV_2015} showed the supremacy of deep learning strategies also in this field. Then, other works improved accuracy by stacking more networks \cite{Ilg_CVPR_2017} or exploiting traditional pyramidal \cite{Ranjan_CVPR_2017,Sun_CVPR_2018, Hui_CVPR_2018} and multi-frame fusion \cite{ren2019fusion} approaches. Unfortunately, obtaining even sparse labels for optical flow is extremely challenging, which renders self-supervision from images highly desirable. For this reason, an increasing number of methods propose to use image reconstruction and spatial smoothness \cite{Jason_ECCV_2016,Ren_AAAI_2017,Guan_ICME_2019} as main signals to guide the training, paying particular attention to occluded regions \cite{Meister_AAAI_2018,Wang_Yang_CVPR_2018, Liu_AAAI_2019, Liu_CVPR_2019, Janai_ECCV_2018,hur2019iterative}.         

\textbf{Semantic segmentation and depth estimation.}
Monocular depth estimation is tightly connected to the semantics of the scene. We can infer the depth of a scene by a single image mostly because of context and prior semantic knowledge. Prior works explored the possibility to learn both tasks with either full supervision \cite{Wang_CVPR_2015, Eigen_ICCV_2015, Mousavian_3DV_2016, kendall2018multi, Zhang_ECCV_2018, Jiao_ECCV_2018, Dovesi_ICRA_2020} or  supervision concerned with semantic labels only 
\cite{Ramirez_ACCV_2018,Chen_CVPR_2019}.
Unlike previous works, we propose a compact architecture trained by self-supervision on monocular videos and exploiting proxy semantic labels.

\textbf{Semantic segmentation and optical flow.}
Joint learning of semantic segmentation and optical flow estimation has been already explored \cite{hur2016joint}. Moreover, scene segmentation \cite{sevilla2016optical,bai2016exploiting} is required to disentangle  potentially moving and static objects for focused optimizations. Differently,  \cite{rashed2019optical} leverages on optical flow to improve semantic predictions of moving objects. Peculiarly w.r.t. previous work, our proposal features a novel self-distillation training procedure guided by semantics to improve occlusion handling.

\textbf{Scene understanding from stereo videos.} Finally, we mention recent works approaching stereo depth estimation with optical flow \cite{aleotti2020learning} and semantic segmentation \cite{Jiang_2019_ICCV} for comprehensive scene understanding. In contrast, we are the first to rely on monocular videos to this aim.


\begin{figure}[t]
\centering
\includegraphics [width=0.50\textwidth]{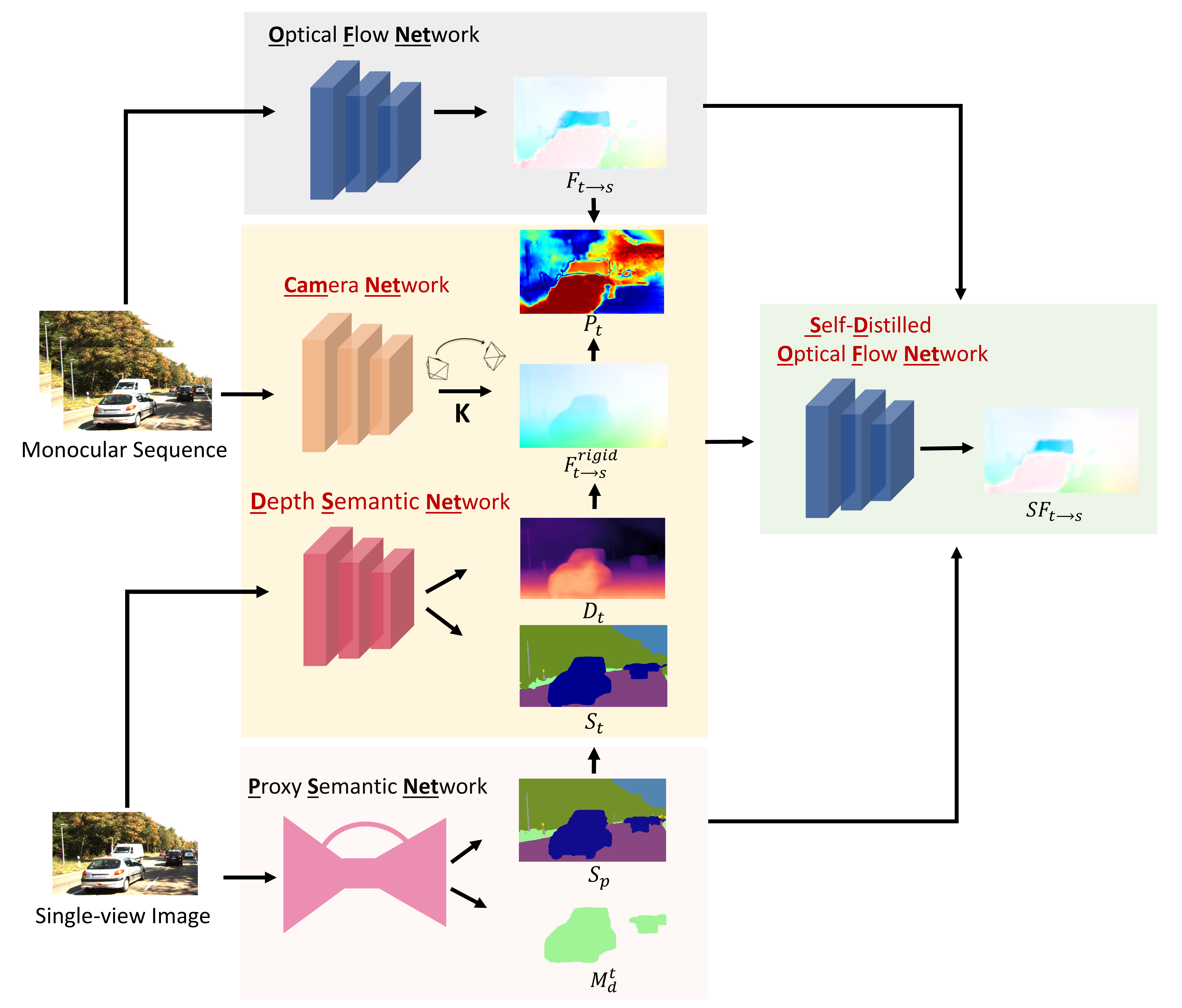}   
\caption{Overall framework for training \netname{} to predict depth, camera pose, camera intrinsics, semantic labels and optical flow. In \textcolor{red}{\textbf{red}} architectures composing \netname{}.}\label{omeganet}
\end{figure}

\section{Overall Learning Framework}

\label{sec:method}
Our goal is to develop a real-time comprehensive scene understanding framework capable of learning strictly related tasks from monocular videos. Purposely, we propose a multi-stage approach to learn first geometry and semantics, then elicit motion information, as depicted in Figure \ref{omeganet}.

\subsection{Geometry and Semantics}

\textbf{Self-supervised depth and pose estimation.} 
We propose to solve a self-supervised single-image depth and pose estimation problem by exploiting geometrical constraints in a sequence of $N$ images, in which one of the frames is used as the target view $I_t$ and the other ones in turn as the source image $I_s$.
Assuming a moving camera in a stationary scene, given a depth map $D_t$ aligned with $I_t$, the camera intrinsic parameters $K$ and the relative pose $T_{t \rightarrow s}$ between $I_{t}$ and $I_s$, it is possible to sample pixels from $I_s$ in order to synthesise a warped image $\widetilde{I}_t$ aligned with $I_t$.
The mapping between corresponding homogeneous pixels coordinates $p_t \in I_t$ and $p_s \in I_s$ is given by:
\begin{equation}\label{eq:rigid_geometry}
    p_{s} \sim K T_{t \rightarrow s} D_{p_{t}} K^{-1}p_{t}
\end{equation}

Following \cite{Zhou_2017_CVPR}, we use the sub-differentiable bilinear sampler mechanism proposed in \cite{Jaderberg_INPS_2015} to obtain $\widetilde{I}_t$.
Thus, in order to learn depth, pose and camera intrinsics  we train two separate CNNs to minimize the photometric reconstruction error between $\widetilde{I}_t$ and $I_{t}$, defined as:

\begin{equation}
\mathcal{L}_{ap}^D = \sum_{p} \psi(I_{t}(p),\widetilde{I_t}(p))
\label{eq:reconstruction_loss}
\end{equation}

where $\psi$ is a photometric error function between the two images.
However, as pointed out in \cite{Godard_ICCV_2019}, such a formulation is prone to errors at occlusion/disocclusion regions or in static camera scenarios. To soften these issues, we follow the same principles as suggested in \cite{Godard_ICCV_2019}, where a minimum per-pixel reprojection loss is used to compute the photometric error, an automask method allows for filtering-out spurious gradients when the static camera assumption is violated, and an edge-aware smoothness loss term is used as in \cite{Godard_CVPR_2017}. Moreover, we use the depth normalization strategy proposed in \cite{Wang_CVPR_2018}. See supplementary material for further details.

We compute the rigid flow between $I_t$ and $I_s$ as the difference between the projected and original pixel coordinates in the target image:

\begin{equation}
 F^{rigid}_{t \rightarrow s} (p_t) = p_{s} - p_t
\label{eq:rigid_flow}
\end{equation}

\textbf{Distilling semantic knowledge.} The proposed distillation scheme is motivated by how time-consuming and cumbersome obtaining accurate pixel-wise semantic annotations is. Thus, we train our framework to estimate semantic segmentation masks $S_t$ by means of supervision from cheap proxy labels $S_p$ distilled by a semantic segmentation network, pre-trained on few annotated samples and capable to generalize well to diverse datasets. Availability of proxy semantic labels for the frames of a monocular video enables us to train a single network to predict jointly depth and semantic labels. Accordingly, the joint loss is obtained  by adding a standard cross-entropy term $\mathcal{L}_{sem}$ to the previously defined self-supervised image reconstruction loss $\mathcal{L}_{ap}^D$.  Moreover, similarly to \cite{Ramirez_ACCV_2018}, we deploy a cross-task loss term, $\mathcal{L}_{edge}^{D}$ (see supplementary), aimed at favouring spatial coherence between depth edges and semantic boundaries. However, unlike  \cite{Ramirez_ACCV_2018}, we do not exploit stereo pairs at training time.

\subsection{Optical Flow and Motion Segmentation}
\label{sec:OFMSmethod}

\textbf{Self-supervised optical flow.} As  the 3D structure of a scene includes stationary as well as non-stationary objects,  to handle the latter we rely on a classical optical flow formulation. Formally, given  two images $I_t$ and $I_s$, the goal is to estimate the 2D motion vectors $F_{t \rightarrow s} (p_t)$ that map each pixel in $I_t$ into its corresponding one in $I_s$. To learn such a mapping without supervision, previous approaches \cite{Meister_AAAI_2018,Liu_CVPR_2019,Yin_CVPR_2018} employ an image reconstruction loss $\mathcal{L}_{ap}^F$ that minimizes the photometric differences between $I_t$ and the back-warped image $\widetilde{I_t}$ obtained by sampling pixels from $I_s$ using the estimated 2D optical flow $F_{t \rightarrow s} (p_t)$. This approach performs well for non-occluded pixels but provides misleading information within occluded regions.

\textbf{Pixel-wise motion probability.}
Non-stationary objects produce systematic errors when optimizing $\mathcal{L}_{ap}^D$  due to the assumption that the camera is the only moving body in an otherwise stationary scene.
However, such systematic errors can be exploited to identify non-stationary objects: at pixels belonging to such objects the rigid flow $F_{t \rightarrow s}^{rigid}$  and the optical flow $F_{t \rightarrow s}$ should exhibit different directions and/or norms. Therefore, a pixel-wise probability of belonging to an object independently moving between frames $s$ and $t$, $P_t$, can be obtained by normalizing the differences between the two vectors. Formally, denoting with $\theta (p_t)$ the angle between the two vectors at location $p_t$, we define the per-pixel motion probabilities as:
\newcommand{\Ffull}{F_{t \rightarrow s} (p_t)}
\newcommand{\Frigid}{F^{rigid}_{t \rightarrow s} (p_t)}
\newcommand{\ltwonorm}[1]{\| #1 \|_2}
\begin{align}
\label{eq:motion_probability}
P_t (p_t) & = \max \{ \frac{1 - \cos \theta (p_t)}{2}, 1 - \rho (p_t) \} 
\end{align}
where $\cos \theta (p_t)$ can be computed as the normalized dot product between the vectors and evaluates the similarity in direction between them, while $\rho(p_t)$ is defined as
\begin{align}
\label{eq:ratio_similarity}
\rho (p_t) & = \frac{\min\{ \ltwonorm{\Ffull} , \ltwonorm{\Frigid} \} }{ \max\{ \ltwonorm{\Ffull} , \ltwonorm{\Frigid} \} } \mbox{,}
\end{align}
\ie a normalized score of the similarity between the two norms. By taking the maximum of the two normalized differences, we can detect moving objects even when either the directions or the norms of the vectors are similar. A visualization of $P_t (p_t)$ is depicted in Fig. \ref{fig:method}(d).

\textbf{Semantic-aware Self-Distillation Paradigm}. Finally, we combine semantic information, estimated optical flow, rigid flow and pixel-wise motion probabilities within a final training stage to obtain a more robust self-distilled optical flow network. In other words, we train a new instance of the model to infer a self-distilled flow SF$_{t \rightarrow s}$ given the estimates F$_{t \rightarrow s}$ from a first self-supervised network and the aforementioned cues.
As previously discussed and highlighted in Figure \ref{fig:method}(c), standard self-supervised  optical flow is prone to errors in occluded regions due to the lack of photometric information  but can provide good estimates for the dynamic objects in the scene. On the contrary, the estimated rigid flow can  properly handle occluded areas thanks to the minimum-reprojection mechanism \cite{Godard_ICCV_2019}. Starting from these considerations, our key idea is to split the scene into stationary and potentially dynamics objects, and apply on them the proper supervision.  
Purposely, we can leverage several observations:

\begin{enumerate}
    \item \textbf{Semantic priors.} Given a semantic map $S_t$ for image $I_t$, we can binarize pixels into static  $M_t^s$ and potentially dynamic $M_t^d$, with $M_t^s \cap M_t^d = \emptyset$. For example, we expect that points labeled as  \textit{road}  are static in the 3D world, while pixels belonging to the semantic class \textit{car} may move. In $M_t^d$, we assign 1 for each potentially dynamic pixel, 0 otherwise, as shown in Figure \ref{fig:method}(e).
    
    \item \textbf{Camera Motion Boundary Mask.} Instead of using a backward-forward strategy \cite{Zou_ECCV_2018} to detect  boundaries occluded due to the ego-motion, we analytically compute a binary boundary mask $M_t^b$ from depth and ego-motion estimates as proposed in \cite{Mahjourian_CVPR_2018}. We assign a 0 value for out-of-camera pixels, 1 otherwise as shown in Figure \ref{fig:method}(f).    
    
    \item \textbf{Consistency Mask.}  Because the inconsistencies between the rigid flow and F$_{t \rightarrow s}$ are not only due to dynamic objects but also to occluded/inconsistent areas, we can leverage Equation \eqref{eq:motion_probability} to detect such critical regions. Indeed, we define the consistency mask as:
        \begin{align}
        \label{eq:inconsistency_mask}
        M_t^{c} & = P_t < \xi,  \xi \in [0,1]
        \end{align}
    This mask assigns 1 where the condition is satisfied, 0 otherwise (\ie{} inconsistent regions) as in Figure \ref{fig:method}(g). 
    
\end{enumerate}{}
Finally, we compute the final mask $M$, in Figure \ref{fig:method}(h), as:

\begin{figure*}[t]
\centering
 \begin{tabular}{c}
    \begin{overpic}[width=0.24\textwidth]{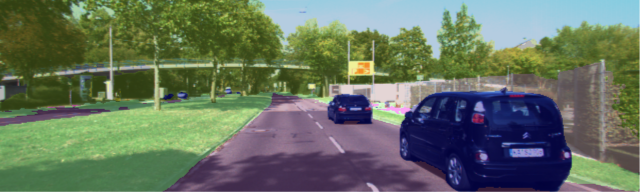}
    \put (0,3) {\colorbox{white}{$\displaystyle\textcolor{black}{\text{(a)}}$}}
    \end{overpic} 
    \begin{overpic}[width=0.24\textwidth]{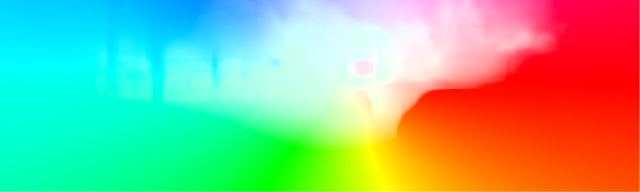}
    \put (0,3) {\colorbox{white}{$\displaystyle\textcolor{black}{\text{(b)}}$}}
    \end{overpic} 
    \begin{overpic}[width=0.24\textwidth]{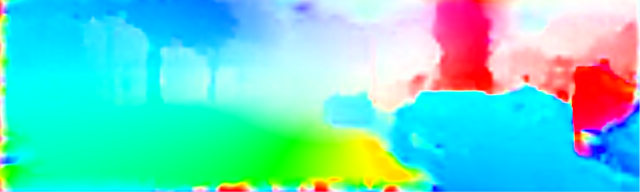}
    \put (0,3) {\colorbox{white}{$\displaystyle\textcolor{black}{\text{(c)}}$}}
    \end{overpic} 
    \begin{overpic}[width=0.24\textwidth]{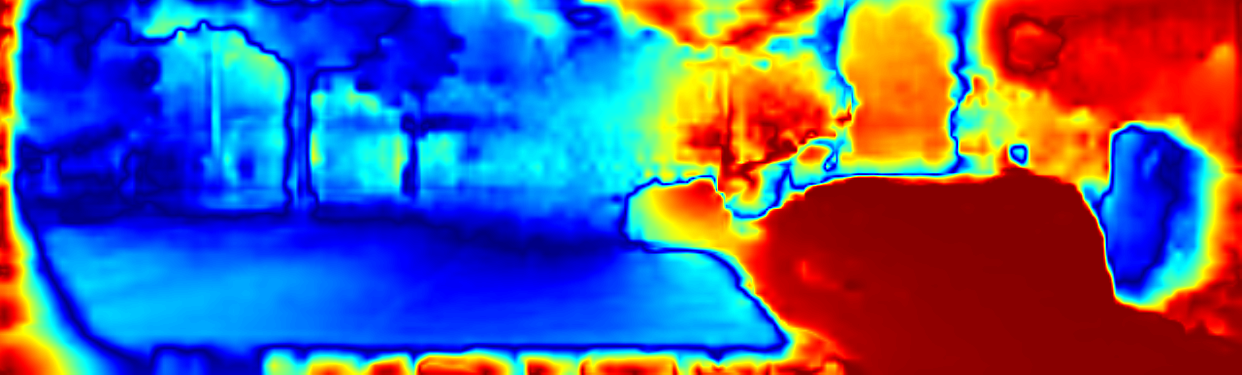}
    \put (0,3) {\colorbox{white}{$\displaystyle\textcolor{black}{\text{(d)}}$}}
    \end{overpic}  
    \\
    \begin{overpic}[width=0.24\textwidth]{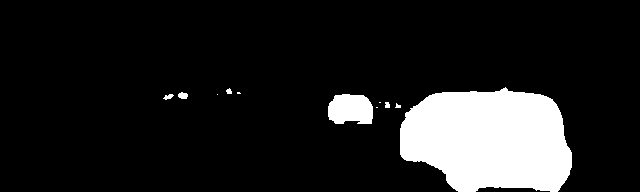}
    \put (0,3) {\colorbox{white}{$\displaystyle\textcolor{black}{\text{(e)}}$}}
    \end{overpic} 
    \begin{overpic}[width=0.24\textwidth]{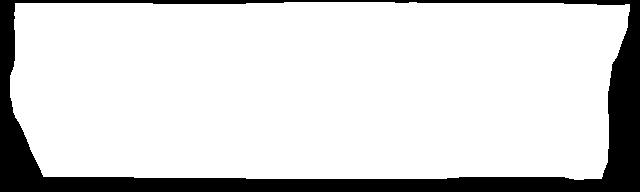}
    \put (0,3) {\colorbox{white}{$\displaystyle\textcolor{black}{\text{(f)}}$}}
    \end{overpic} 
    \begin{overpic}[width=0.24\textwidth]{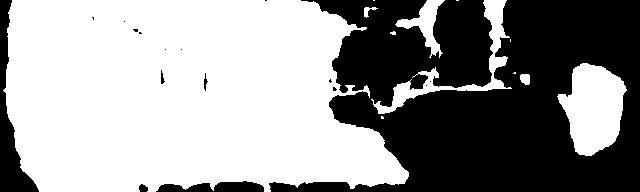}
    \put (0,3) {\colorbox{white}{$\displaystyle\textcolor{black}{\text{(g)}}$}}
    \end{overpic} 
    \begin{overpic}[width=0.24\textwidth]{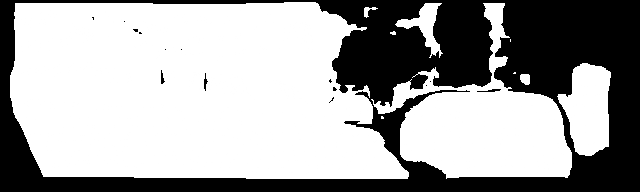}
    \put (0,3) {\colorbox{white}{$\displaystyle\textcolor{black}{\text{(h)}}$}}
    \end{overpic} 
\end{tabular}
\vspace{1mm}
\caption{Overview of our semantic-aware and self-distilled optical flow estimation approach. We leverage semantic segmentation $S_t$ (a) together with rigid flow $F^{rigid}_{t\rightarrow s}$ (b), teacher flow $F_{t\rightarrow s}$ (c) and motion probabilities $P_t$ (d), the warmer the higher. From a) we obtain semantic priors $M^d_t$ (e), combined with boundary mask $M^b_t$ (f) and consistency mask $M^c_t$ (g) derived from (d) as in Eq. \ref{eq:inconsistency_mask}, in order to obtain the final mask $M$ (h) as in Eq. \ref{eq:final_mask}. }
\label{fig:method}

\end{figure*}

\begin{align}
\label{eq:final_mask}
M = \min\{\max\{M_t^d, M_t^c \}, M_t^b \}
\end{align}

As a consequence, $M$ will effectively distinguish regions in the image for which we can not trust the supervision sourced by $F_{t \rightarrow s}$, \ie inconsistent or occluded areas. On such regions, we can leverage our proposed self-distillation mechanism.
Then, we define the final total loss for the self-distilled optical flow network as:
\begin{multline}
 \mathcal{L} = \sum \alpha_r \phi(SF_{t \rightarrow s} ,F_{t \rightarrow s}^{rigid}) \cdot (1 - M) \\ + \alpha_d \phi(SF_{t \rightarrow s} ,F_{t \rightarrow s}) \cdot M + \psi(I_{t},\widetilde{I}_{t}^{SF}) \cdot M
 \label{eq:lsflow}
\end{multline}

where $\phi$ is a distance function between two motion vectors, while $\alpha_r$ and $\alpha_d$ are two hyper-parameters.

\subsection{Motion Segmentation}
At test time, from pixel-wise probability $P_t$ computed between $SF_{t \rightarrow s}$ and $F_{t \rightarrow s}^{rigid}$, semantic prior $M_t^d$ and a threshold $\tau$, we compute a motion segmentation mask by:

\begin{align}
\label{eq:motion_mask}
M_t^{mot} = M_t^d \cdot  (P_t > \tau), \tau \in [0,1]
\end{align}

Such mask allows us to detect moving objects in the scene independently of the camera motion. A qualitative example is depicted in Figure \ref{fig:abstract}(f).


\section{Architecture and Training Schedule} 
In this section we present the networks composing \netname{} (highlighted in \textcolor{red}{red} in Figure \ref{omeganet}), and delineate their training protocol. We set $N=3$, using 3-frames sequences.
The source code is available at \url{https://github.com/CVLAB-Unibo/omeganet}.

\subsection{Network architectures}

We highlight the key traits of each network, referring the reader to the supplementary material for exhaustive details.

\textbf{\underline{D}epth and \underline{S}emantic \underline{Net}work (DSNet).}
We build a single model, since shared reasoning about the two tasks is beneficial to both \cite{Ramirez_ACCV_2018,Chen_CVPR_2019}.
To achieve real-time performance, DSNet is inspired to PydNet \cite{Poggi_IROS_2018}, with several key modifications due to the different goals. We extract a pyramid of features down to $\frac{1}{32}$ resolution, estimating a first depth map at the bottom. Then, it is upsampled and concatenated with higher level features in order to build a refined depth map. We repeat this procedure up to half resolution, where two estimators predict the final depth map $D_t$ and semantic labels $S_t$. These are bi-linearly upsampled to full resolution. Each conv layer is followed by batch normalization and ReLU, but the prediction layers, using reflection padding. DSNet counts $1.93$M parameters. 

\textbf{\underline{Cam}era \underline{Net}work (CamNet).}
This network estimates both camera intrinsics and poses between a target $I_t$ and some source views $I_s ( 1 \leq s \leq 3, s \neq t)$.
CamNet differs from previous work by extracting features from $I_t$ and $I_s$ independently with shared encoders. We extract a pyramid of features down to $\frac{1}{16}$ resolution for each image and concatenate them to estimate the 3 Euler angles and the 3D translation for each $I_s$. As in \cite{Gordon_ICCV_2019}, we also estimate the camera intrinsics. Akin to DSNet, we use batch normalization and ReLU after each layer but for prediction layers. CamNet requires  $1.77$M parameters for pose estimation and $1.02$K for the camera intrinsics.

\textbf{\underline{O}ptical \underline{F}low \underline{Net}work (OFNet).} 
To pursue real-time performance, we deploy a 3-frame PWC-Net \cite{Sun_CVPR_2018} network as in \cite{Liu_CVPR_2019}, which counts $4.79$M parameters. Thanks to our novel training protocol leveraging on semantics and self-distillation, our OFNet can outperform other multi-task frameworks \cite{Ranjan_CVPR_2019} built on the same optical flow architecture. 

\subsection{Training Protocol}
Similarly to \cite{Yin_CVPR_2018}, we employ a two stage learning process to facilitate the network optimisation process. At first, we train DSNet and CamNet simultaneously, then we train OFNet by the self-distillation paradigm described in \ref{sec:OFMSmethod}. For both stages, we use a batch size of 4 and resize input images to $640 \times 192$ for the KITTI dataset (and to $768 \times 384$ for  pre-training on Cityscapes), optimizing the output of the networks at the highest resolution only. We also report additional experimental results for different input resolutions where specified. We use the Adam optimizer \cite{Kingma_2014} with $\beta_{1} = 0.9$, $\beta_{2} = 0.999$ and $\epsilon=10^{-8}$. As photometric loss $\psi$, we employ the same function defined in \cite{Godard_CVPR_2017}. When training our networks, we apply losses using as $I_s$ both the previous and the next image of our 3-frame sequence. Finally, we set both $\tau$ and $\xi$ to be $0.5$ in our experiments.

\textbf{Depth, Pose, Intrinsics and Semantic Segmentation.}
In order to train DSNet and CamNet we employ sequences of 3 consecutive frames and semantic proxy labels yielded by a state-of-the art architecture \cite{Chen_NIPS_2018} trained on Cityscapes with ground-truth labels.
We trained DSNet and CamNet for $300$K iterations, setting the initial learning rate to $10^{-4}$, manually halved after $200$K, $250$K and $275$K steps. We apply data augmentation to images as in \cite{Godard_CVPR_2017}. Training takes $\sim20$ hours on a Titan Xp GPU.


\begin{table*}[t]
\centering
\scalebox{0.76}{
\begin{tabular}{l|cc|cc|cccc|ccc}
\multicolumn{7}{c}{} &\multicolumn{2}{c}{\cellcolor{blue!25} Lower is better}
 & \multicolumn{2}{c}{\cellcolor{LightCyan} Higher is better} \\
\hline
Method & M  & A & I & CS & \cellcolor{blue!25} Abs Rel & \cellcolor{blue!25} Sq Rel & \cellcolor{blue!25} RMSE & \cellcolor{blue!25} RMSE log &  \cellcolor{LightCyan}$\delta<$1.25 &  \cellcolor{LightCyan}$\delta<1.25^2$ & \cellcolor{LightCyan}$\delta<1.25^3$ \\
\hline

\hline

Godard \etal \cite{Godard_ICCV_2019} & & & & & 0.132 & 1.044 & 5.142 & 0.210 & 0.845 & 0.948 & 0.977 \\
Godard \etal \cite{Godard_ICCV_2019} ($1024 \times 320$) & & & \checkmark &  & \textbf{0.115} & 0.882 & 4.701 & 0.190 & \textbf{0.879} & \textbf{0.961} & 0.982 \\
Zhou \etal \cite{zhou2019unsupervised} & & & \checkmark &  & 0.121 & 0.837 & 4.945 & 0.197 & 0.853 & 0.955 & 0.982 \\
Mahjourian \etal \cite{Mahjourian_CVPR_2018} &  & & & \checkmark & 0.159 & 1.231 & 5.912 & 0.243 & 0.784 & 0.923 & 0.970 \\
Wang \etal \cite{Wang_CVPR_2018} & & & & \checkmark & 0.151 & 1.257 & 5.583 & 0.228 & 0.810 & 0.936 & 0.974 \\
Bian \etal \cite{Bian_NeurIPS_2019} & &  & & \checkmark & 0.128 & 1.047 & 5.234 & 0.208 & 0.846 & 0.947 & 0.970 \\
Yin \etal \cite{Yin_CVPR_2018} & \checkmark & & & \checkmark & 0.153 & 1.328 & 5.737 & 0.232 & 0.802 & 0.934 & 0.972 \\
Zou \etal \cite{Zou_ECCV_2018} & \checkmark & & & \checkmark & 0.146 & 1.182 & 5.215 & 0.213 & 0.818 & 0.943 & 0.978 \\
Chen \etal \cite{Chen_ICCV_2019} & \checkmark & & \checkmark & & 0.135 & 1.070 & 5.230 & 0.210 & 0.841 & 0.948 & 0.980 \\
Luo \etal \cite{Luo_EPC++_2018} & \checkmark & & & & 0.141 & 1.029 & 5.350 & 0.216 & 0.816 & 0.941 & 0.976 \\
Ranjan \etal \cite{Ranjan_CVPR_2019} & \checkmark & & & & 0.139 & 1.032 & 5.199 & 0.213 & 0.827 & 0.943 & 0.977 \\
Xu \etal \cite{Xu_IJCAI_2019} & & \checkmark & \checkmark & & 0.138 & 1.016 & 5.352 & 0.217 & 0.823 & 0.943 & 0.976 \\
Casser \etal \cite{Casser_AAAI_2019} & & \checkmark & & & 0.141 & 1.026 & 5.290 & 0.215 & 0.816 & 0.945 & 0.979 \\
Gordon \etal \cite{Gordon_ICCV_2019} & \checkmark & \checkmark  & & & 0.128 & 0.959 & 5.230 & - & - & - & - \\
\hline
\textbf{\netname ($640 \times 192$)} & \checkmark & \checkmark & & & 0.126 & 0.835 & 4.937 & 0.199 & 0.844 & 0.953 & 0.982 \\
\textbf{\netname ($1024 \times 320$)} & \checkmark & \checkmark & & & 0.125 & 0.805 & 4.795 & 0.195 & 0.849 & 0.955 & 0.983 \\
\textbf{\netname ($640 \times 192$)} & \checkmark & \checkmark & & \checkmark & 0.120 & 0.792 & 4.750 & 0.191 & 0.856 & 0.958 & 0.984 \\
\textbf{\netname ($1024 \times 320$)} & \checkmark & \checkmark & & \checkmark & 0.118 & \textbf{0.748} & \textbf{4.608} & \textbf{0.186} & 0.865 & \textbf{0.961} & \textbf{0.985} \\
\hline
\end{tabular}
}
\smallskip
\caption{Depth evaluation on the Eigen split \cite{Eigen_2014} of KITTI \cite{Geiger_SAGE_2013}. We indicate additional features of each method. M: multi-task learning, A: additional information (\eg object knowledge, semantic information), I: feature extractors pre-trained on ImageNet \cite{Deng_IEEE_2009}, CS:  network pre-trained on Cityscapes \cite{Cordts_CVPR_2016}.} 
\label{table:eigen}
\end{table*}

\textbf{Optical Flow.}
We train OFNet by the procedure presented in \ref{sec:OFMSmethod}. In particular, we perform $200$K training steps with an initial learning rate of $10^{-4}$, halved every $50$K until convergence. 
Moreover, we apply strong data augmentation consisting in random horizontal and vertical flip, crops, random time order switch and, peculiarly, time stop, replacing all $I_s$ with $I_t$ to learn a zero motion vector. This configuration requires about $13$ hours on a Titan Xp GPU with the standard $640 \times 192$ resolution. We use an L1 loss as $\phi$.
Once obtained a competitive network in non-occluded regions we train a more robust optical flow network, denoted as SD-OFNet, starting from pre-learned weights and the same structure of OFNet by distilling knowledge from OFNet and rigid flow computed by DSNet using the total mask $M$ and $416 \times 128$ random crops applied to $F_{t \rightarrow s}$, $F^{rigid}_{t \rightarrow s}$, $M$ and RGB images. We train SD-OFNet for $15K$ steps only with a learning rate of $2.5\times10^{-5}$ halved after $5$K, $7.5$K, $10$K and $12.5$K steps, setting $\alpha_r$ to 0.025 and $\alpha_d$ to 0.2. At test-time, we rely on SD-OFNet only. 


\section{Experimental results}
\label{sec:experiments}

Using standard benchmark datasets, we present here the experimental validation on the main tasks tackled by \netname{}.

\subsection{Datasets.}

We conduct experiments on standard benchmarks such as KITTI and Cityscapes. 
We do not use feature extractors pre-trained on ImageNet or other datasets. For the sake of space, we report further studies in the supplementary material (\eg results on pose estimation or generalization).

\textbf{KITTI} (K) \cite{Geiger_CVPR_2012} is a collection of 42,382 stereo sequences taken in urban  environments from two video cameras and a LiDAR device mounted on the roof of a car. This dataset is widely used for benchmarking geometric understanding tasks such as depth, flow and pose estimation.  

\textbf{Cityscapes} (CS) \cite{Cordts_CVPR_2016} is an outdoor dataset containing stereo pairs taken from a moving vehicle in various weather conditions. This dataset features higher resolution and higher quality images. While sharing  similar settings, this dataset contains more dynamics scenes compared to KITTI.  It consists of 22,973 stereo pairs with $2048 \times 1024$ resolution. 2,975 and 500 images come with fine semantic 

\begin{table*}[!htbp]
\centering
\scalebox{0.64}{
\begin{tabular}{l|cccccc|cccc|ccc}
\multicolumn{9}{c}{} &\multicolumn{2}{c}{\cellcolor{blue!25} Lower is better}
 & \multicolumn{2}{c}{\cellcolor{LightCyan} Higher is better} \\
\hline
Resolution & Learned Intr. \cite{Gordon_ICCV_2019} &  Norm. \cite{Wang_CVPR_2018} & Min. Repr. \cite{Godard_ICCV_2019} & Automask \cite{Godard_ICCV_2019} & Sem. \cite{Chen_NIPS_2018} & Pre-train & \cellcolor{blue!25} Abs Rel & \cellcolor{blue!25} Sq Rel & \cellcolor{blue!25} RMSE & \cellcolor{blue!25} RMSE log &  \cellcolor{LightCyan}$\delta<$1.25 &  \cellcolor{LightCyan}$\delta<1.25^2$ & \cellcolor{LightCyan}$\delta<1.25^3$ \\
\hline

\hline
\textbf{$640 \times 192$} & - & - & - & - & - & - & 0.139 & 1.056 & 5.288 & 0.215 & 0.826 & 0.942 & 0.976 \\
\textbf{$640 \times 192$} & \checkmark & - & - & - & - & - & 0.138 & 1.014 & 5.213 & 0.213 & 0.829 & 0.943 & 0.977 \\
\textbf{$640 \times 192$} & \checkmark & \checkmark & - & - & - & - &  0.136 & 1.008 & 5.204 & 0.212 & 0.832 & 0.944 & 0.976 \\
\textbf{$640 \times 192$} & \checkmark & \checkmark & \checkmark & - & - & - & 0.132 & 0.960 & 5.104 & 0.206 & 0.840 & 0.949 & 0.979 \\
\textbf{$640 \times 192$} & \checkmark & \checkmark & \checkmark & \checkmark & - & - & 0.130 & 0.909 & 5.022 & 0.207 & 0.842 & 0.948 & 0.979 \\
\textbf{$640 \times 192$} \textdagger & \checkmark & \checkmark & \checkmark & \checkmark & - & - & 0.134 & 1.074 & 5.451 & 0.213 & 0.834 & 0.946 & 0.977 \\
\textbf{$640 \times 192$} & \checkmark & \checkmark & \checkmark & \checkmark & \checkmark & - & 0.126 & 0.835 & 4.937 & 0.199 & 0.844 & 0.953 & 0.980 \\
\hline
\textbf{$416 \times 128$} & \checkmark & \checkmark & \checkmark & \checkmark & \checkmark & \checkmark & 0.126 & 0.862 & 4.963 & 0.199 & 0.846 & 0.952 & 0.981 \\
\textbf{$640 \times 192$} & \checkmark & \checkmark & \checkmark & \checkmark & \checkmark & \checkmark & 0.120 & 0.792 & 4.750 & 0.191 & 0.856 & 0.958 & 0.984 \\
\textbf{$1024 \times 320$} & \checkmark & \checkmark & \checkmark & \checkmark & \checkmark & \checkmark & 0.118 & 0.748 & 4.608 & 0.186 & 0.865 & 0.961 & 0.985 \\

\hline
\end{tabular}
}
\smallskip
\caption{Ablation study of our depth network on the Eigen split \cite{Eigen_2014} of KITTI. \textdagger{}: our network is replaced by a ResNet50 backbone \cite{Yin_CVPR_2018}. 
}
\label{table:ablation}
\end{table*}

\subsection{Monocular Depth Estimation}

In this section, we  compare our results to other state-of-the-art proposals and assess the contribution of each component to the quality of our monocular depth predictions.

\textbf{Comparison with state-of-the-art}. 
We compare with state-of-the-art self-supervised networks trained on monocular videos according to the protocol described in \cite{Eigen_2014}.
We follow the same pre-processing procedure as \cite{Zhou_2017_CVPR} to remove static images from the training split while using all the 697 images for testing. LiDAR points  provided in \cite{Geiger_CVPR_2012} are reprojected on the left input image to obtain ground-truth labels for evaluation, up to 80 meters \cite{Garg_ECCV_2016}. Since the predicted depth is defined up to a scale factor, we align the scale of our estimates by multiplying them by a scalar that matches the median of the ground-truth, as introduced in \cite{Zhou_2017_CVPR}. We adopt the standard performance metrics defined in \cite{Eigen_2014}. Table \ref{table:eigen} reports extensive comparison with respect to several monocular depth estimation methods. 
We outperform our main competitors such as \cite{Yin_CVPR_2018, Zou_ECCV_2018, Chen_ICCV_2019, Ranjan_CVPR_2019} that solve multi-task learning or other strategies that exploit additional information during the training/testing phase \cite{Casser_AAAI_2019, Xu_IJCAI_2019}. Moreover, our best configuration, \ie pre-training on CS and using $1024 \times 320$ resolution, achieves better results in 5 out of 7 metrics with respect to the single-task, state-of-the-art proposal \cite{Godard_ICCV_2019} (and is the second best and very close to it on the remaining 2) which, however, leverages on a larger ImageNet pre-trained model based on ResNet-18. It is also interesting to note how our proposal without pretraining obtains the best performance in 6 out of 7 measures on $640 \times 192$ images (row 1 vs 15). These results validate our intuition about how the use of semantic information can guide geometric reasoning and make a compact network provide state-of-the-art performance even with respect to larger and highly specialized depth-from-mono methods.

\begin{table}
\centering
\scalebox{0.7}{
\begin{tabular}{l|c|cccc}
\hline
Method & Cap (m) & \cellcolor{blue!25} Abs Rel & \cellcolor{blue!25} Sq Rel & \cellcolor{blue!25} RMSE & \cellcolor{blue!25} RMSE log\\

\hline
Godard \etal \cite{Godard_ICCV_2019} & 0-8 & \textbf{0.059} & \textbf{0.062} & 0.503 & \textbf{0.082}\\
\textbf{\netname}\textdagger  & 0-8 & 0.060 & 0.063 & \textbf{0.502} & \textbf{0.082} \\
\textbf{\netname} & 0-8 & 0.062 & 0.065 & 0.517 &  0.085\\

\hline
Godard \etal \cite{Godard_ICCV_2019} & 0-50 & 0.125 & 0.788 & 3.946 & 0.198\\
\textbf{\netname}\textdagger  & 0-50 & 0.127 & 0.762 & 4.020 & 0.199 \\
\textbf{\netname} & 0-50 & \textbf{0.124} & \textbf{0.702} & \textbf{3.836} & \textbf{0.195}\\

\hline
Godard \etal \cite{Godard_ICCV_2019} & 0-80 & 0.132 & 1.044 & 5.142 & 0.210\\
\textbf{\netname}\textdagger  & 0-80 & 0.134 & 1.074 & 5.451 & 0.213 \\
\textbf{\netname} & 0-80 & \textbf{0.126} & \textbf{0.835} & \textbf{4.937} & \textbf{0.199}\\

\hline
\end{tabular}
}
\smallskip
\caption{Depth errors by varying the range. \textdagger{}: our network is replaced by a ResNet50 backbone \cite{Yin_CVPR_2018}. }
\label{table:depth_caps_main}
\end{table}

\textbf{Ablation study}. Table \ref{table:ablation} highlights how progressively adding the key innovations proposed in \cite{Gordon_ICCV_2019, Godard_ICCV_2019, Wang_CVPR_2018} contributes to strengthen \netname{}, already comparable to other methodologies even in its baseline configuration (first row).
Interestingly, a large improvement is achieved by deploying joint depth and semantic learning (rows 5 vs 7), which forces the network to simultaneously reason about geometry and content within the same shared features.
By replacing DSNet within \netname{} with a larger backbone \cite{Yin_CVPR_2018} (rows 5 vs 6) we obtain worse performance, validating the design decisions behind our compact model.
Finally, by pre-training on CS we achieve the best accuracy, which increases alongside with the input resolution (rows 8 to 10). 

\textbf{Depth Range Error Analysis.} We dig into our depth evaluation to explain the effectiveness of \netname{} with respect to much larger networks. Table \ref{table:depth_caps_main} compares, at different depth ranges, our model with more complex ones \cite{Godard_ICCV_2019,Yin_CVPR_2018}.
This experiment shows how \netname{} superior performance comes from better estimation of large depths: \netname{} outperforms both competitors when we include distances larger than 8 m in the evaluation, while it turns out less effective in the close range. 

\begin{table}[t]
\centering
\scalebox{0.71}{
\begin{tabular}{l|cc|ccc}
\hline
Method & Train & Test & mIoU Class & mIoU Cat. & Pix.Acc.\\
\hline
DABNet \cite{Li_BMVC_2019} & CS(\textcolor{red}{S}) & CS & 69.62 & 87.56 & 94.62\\
FCHardNet \cite{Chao_ICCV_2019} & CS(\textcolor{red}{S}) & CS & \textbf{76.37} & \textbf{89.22} & \textbf{95.35}\\
\textbf{\netname} & CS(\textcolor{YellowOrange}{P}) & CS & 54.80 & 82.92 & 92.50\\
\hline
DABNet \cite{Li_BMVC_2019} & CS(\textcolor{red}{S}) & K & 35.40 & 61.49 & 80.50\\
FCHardNet \cite{Chao_ICCV_2019} & CS(\textcolor{red}{S}) & K & \textbf{44.74} & 68.20 & 72.07\\
\textbf{\netname} & CS(\textcolor{YellowOrange}{P}) & K & 43.80 & \textbf{74.31} & \textbf{88.31}\\
\hline
\textbf{\netname} & CS(\textcolor{YellowOrange}{P}) + K(\textcolor{YellowOrange}{P}) & K & 46.68 & 75.84 &88.12 \\
\hline
\end{tabular}
}
\smallskip
\caption{Semantic segmentation on Cityscapes (CS) and KITTI (K). \textcolor{red}{S}: training on ground-truth, \textcolor{YellowOrange}{P}: training on proxy labels. }
\label{table:semantic_eval}
\end{table}

\subsection{Semantic Segmentation}
In Table \ref{table:semantic_eval}, we report the performance of \netname{} on semantic segmentation for the 19 evaluation classes of CS  according to  the metrics defined in \cite{Cordts_CVPR_2016, Badrinarayanan_PAMI_2017}. We compare \netname{} against state-of-the art networks for real-time semantic segmentation \cite{Chao_ICCV_2019,Li_BMVC_2019} when training on CS and testing either on the validation set of CS (rows 1-3) or the 200 semantically annotated images of K (rows 4-6). 
Even though our network is not as effective as the considered methods when training and testing on the same dataset, it shows greater generalization capabilities to unseen domains: it significantly outperforms other methods when testing on K for mIoU$_\text{category}$ and pixel accuracy, and provides  similar results to \cite{Chao_ICCV_2019} for mIoU$_\text{class}$.
We relate this ability to our training protocol based on proxy labels (\textcolor{YellowOrange}{P}) instead of ground truths (\textcolor{red}{S}). We validate this hypothesis with thorough ablation studies reported in the supplementary material. Moreover, as we have already effectively distilled the knowledge from DPC \cite{Chen_NIPS_2018} during pre-training on CS, there is only a slight benefit in training on both CS and K (with proxy labels only) and testing on K (row 7). Finally, although achieving 46.68 mIoU on fine segmentation, we obtain 89.64 mIoU for the task of segmenting static from potentially dynamic classes, an important result to obtain accurate motion masks.

\begin{table}[t]
\centering
\scalebox{0.72}{
\begin{tabular}{l|c|ccc|c}

\multicolumn{2}{c}{}& \multicolumn{3}{c}{train} & test\\
\hline
Method & Dataset & Noc & All & F1 & F1 \\
\hline
Meister\etal \cite{Meister_AAAI_2018} - C  & SYN + K  & - & 8.80 & 28.94\% & 29.46\% \\
Meister \etal \cite{Meister_AAAI_2018} - CSS  & SYN + K & - & 8.10 & 23.27\% & 23.30\% \\
Zou \etal \cite{Zou_ECCV_2018} & SYN + K  & - & 8.98 & 26.0\% & 25.70\% \\
Ranjan \etal \cite{Ranjan_CVPR_2019} & SYN + K & - & 5.66 & 20.93\% & 25.27\% \\
\hline
Wang \etal \cite{wang2019unos} **& K & - & 5.58  & - & 18.00\%\\
\hline
Yin \etal \cite{Yin_CVPR_2018} & K & 8.05 & 10.81 & - & - \\
Chen \etal \cite{Chen_ICCV_2019} \textdagger & K & 5.40 & 8.95 & - & - \\
Chen \etal \cite{Chen_ICCV_2019} (online)  \textdagger & K & 4.86 & 8.35 & - & - \\
Ranjan \etal \cite{Ranjan_CVPR_2019} & K & - & 6.21 & 26.41\% & - \\
Luo \etal \cite{Luo_EPC++_2018} & K & - & 5.84 & - & 21.56\%  \\
Luo \etal \cite{Luo_EPC++_2018} * & K & - & 5.43 & - & 20.61\%  \\

\textbf{\netname} (Ego-motion) & K  & 11.72 & 13.50 & 51.22\% & - \\
\textbf{OFNet} & K & 3.48 & 11.61 & 25.78\% & - \\
\textbf{SD-OFNet} & K  & \textbf{3.29} & \textbf{5.39} & \textbf{20.0}\%  &\textbf{19.47\%}\\
\hline
\end{tabular}
}
\smallskip
\caption{Optical flow evaluation on the KITTI 2015 dataset. \textdagger{}: pre-trained on ImageNet, SYN: pre-trained on SYNTHIA \cite{RosCVPR16}, *: trained on stereo pairs, **: using stereo at testing time.}
\label{table:kitti2015_of}
\end{table}

\subsection{Optical Flow}
In Table \ref{table:kitti2015_of}, we compare the performance of our optical flow network with competing methods using the KITTI 2015 stereo/flow training set \cite{Geiger_SAGE_2013} as testing set, which contains 200 ground-truth optical flow measurements for evaluation. We exploit all the raw K images for training, but we exclude the images used at testing time as done in  \cite{Zou_ECCV_2018}
, to be consistent with experimental results of previous self-supervised optical flow strategies \cite{Yin_CVPR_2018, Zou_ECCV_2018, Chen_ICCV_2019, Ranjan_CVPR_2019}. From the table, we can observe how our self-distillation strategy allows SD-OFNet to outperform by a large margin competitors trained on K only (rows 5-11), and it even performs better than models pre-initialized by training on synthetic datasets \cite{RosCVPR16}. 
Moreover, we submitted our flow predictions to the online KITTI flow benchmark after retraining the network including images from the whole official training set. In this configuration, we can observe how our model achieves state-of-the-art $F1$ performances with respect to other monocular multi-task architectures.

\subsection{Motion Segmentation}
In Table \ref{table:kitti_motion_segmentation} we report experimental results for the motion segmentation task on the KITTI 2015 dataset, which provides 200 images manually annotated with motion labels for the evaluation. We compare our methodology with respect to other state-of-the-art strategies that performs multi-task learning and motion segmentation \cite{Ranjan_CVPR_2019, Luo_EPC++_2018, wang2019unos} using the metrics and evaluation protocol proposed in \cite{Luo_EPC++_2018}. It can be noticed how our segmentation strategy outperforms all the other existing methodologies by a large margin. This demonstrates the effectiveness of our proposal to jointly combine semantic reasoning and motion probability to obtain much better results. We also report, as upper bound, the accuracy enabled by injecting semantic proxies \cite{Chen_NIPS_2018} in place of \netname{} semantic predictions to highlight the low margin between the two. 

\begin{table}[t]
\centering
\scalebox{0.7}{
\begin{tabular}{l|ccccc}
\hline
Method & Pixel Acc. & Mean Acc. & Mean IoU & f.w. IoU \\
\hline
Yang \etal \cite{Yang_ECCV_Workshops_2018} * & 0.89 & 0.75 & 0.52 & 0.87\\
Luo \etal \cite{Luo_EPC++_2018} & 0.88 & 0.63 & 0.50 & 0.86 \\
Luo \etal \cite{Luo_EPC++_2018} * &0.91 & 0.76 & 0.53 & 0.87\\
Wang \etal \cite{wang2019unos} (Full) **  & 0.90 & 0.82 & 0.56 & 0.88\\
Ranjan \etal \cite{Ranjan_CVPR_2019} & 0.87 & 0.79 & 0.53 & 0.85 \\

\textbf{\netname} & \bfseries 0.98 & \bfseries 0.86 & \bfseries 0.75 & \bfseries 0.97\\
\hline
\textbf{\netname} (Proxy \cite{Chen_NIPS_2018}) & 0.98 & 0.87 & 0.77 & 0.97\\

\hline
\end{tabular}
}
\smallskip
\caption{Motion segmentation evaluation on the KITTI 2015 dataset. *: trained on stereo pairs, **: using stereo at testing time.}
\label{table:kitti_motion_segmentation}
\end{table}


\begin{table}[t]
\centering
\scalebox{0.83}{
\begin{tabular}{l|c|ccccc}
\cline{3-7}
\hline
Device & Watt & D & DS & OF & Cam & $\Omega$ \\
\hline
Jetson TX2 & 15 & 12.5 & 10.3 & 6.5 & 49.2 & 4.5 \\
i7-7700K & 91 & 5.0 & 4.2 & 4.9 & 31.4 & 2.4 \\
Titan XP  & 250 & 170.2 & 134.1 & 94.1 & 446.7 & 57.4 \\
\hline

\end{tabular}
}
\smallskip
\caption{Runtime analysis on different devices. We report the power consumption in Watt and the FPS. D: Depth, S: Semantic, OF: Optical Flow, Cam: camera pose, $\Omega$: Overall architecture. }
\label{table:runtime}
\end{table}

\subsection{Runtime analysis}

Finally, we measure the runtime of \netname{} on different hardware devices, \ie a Titan Xp GPU, an embedded NVIDIA Jetson TX2 board and an Intel i7-7700K@4.2 GHz CPU. Timings averaged over 200 frames at $640 \times 192$ resolution. 
Moreover, as each component of \netname{} may be used on its own, we report the runtime for each independent task. 
As summarized in Table \ref{table:runtime}, our network runs in real-time on the Titan Xp GPU and at about 2.5 FPS on a standard CPU. It also fits the low-power NIVIDA Jetson TX2, achieving 4.5 FPS to compute all the outputs. Additional experiments are available in the supplementary material.


\section{Conclusions}
\label{sec:conclusions}

In this paper, we have proposed the first real-time network for comprehensive scene understanding from monocular videos. Our framework reasons jointly about geometry, motion and semantics in order to estimate accurately depth, optical flow, semantic segmentation and motion masks at about 60 FPS on high-end GPU and 5FPS on embedded systems. To address the above  multi-task problem we have proposed a novel learning procedure based on distillation of proxy semantic labels and semantic-aware self-distillation of optical-flow information. Thanks to this original paradigm, we have demonstrated  state-of-the-art performance on standard benchmark datasets for depth and optical flow estimation as well as for motion segmentation. 

As for future research, we find it intriguing to investigate on whether and how would it be possible to self-adapt \netname{} on-line. Although some very recent works have explored this topic for depth-from-mono \cite{Casser_AAAI_2019} and optical flow \cite{Chen_ICCV_2019}, the key issue with our framework would be to conceive a strategy to deal with semantics. 

\textbf{Acknowledgement.} We gratefully acknowledge the support of NVIDIA Corporation with the donation of the Titan X Pascal GPU used for this research. 

{\small
\bibliographystyle{ieee_fullname}
\bibliography{egbib}
}

\newpage\phantom{Supplementary}
\multido{\i=1+1}{15}{


\section*{Supplementary material}

# Distilled Semantics for Comprehensive Scene Understanding from Videos – Supplementary material

Fabio Tosi[*]  Filippo Aleotti[*]  Pierluigi Zama Ramirez[*]
Matteo Poggi  Samuele Salti  Luigi Di Stefano  Stefano Mattoccia
Department of Computer Science and Engineering (DISI)
University of Bologna, Italy
[*]{fabio.tosi5, filippo.aleotti2, pierluigi.zama}@unibo.it

## Supplementary material

This document provides additional material concerning CVPR 2020 paper, "Distilled Semantics for Comprehensive Scene Understanding from Videos". In particular, we report here a more detailed description of our $\Omega$Net architecture and the losses used to train it, alongside with more insights related to performance in the addressed tasks (depth, pose, optical flow, semantic and motion segmentation) and runtime. Moreover, we include additional qualitative results on KITTI (K) and CityScapes (CS), as well as on an arbitrary YouTube video for which the camera parameters are not known in advance, thus showing how $\Omega$Net can provide comprehensive scene understanding in the wild.

## 1. Network Architecture

In this section, we provide a more detailed description of our $\Omega$Net architecture.

Table 1 reports a detailed specification of the layers building up the DSNet and CamNet modules. For each layer, we report kernel size (K), stride (S) and number of input/output channels. As for OFNet and the proxy semantic network, a thorough description can be found in [17] and [6] respectively.

## 2. Losses

To train the DSNet module, we rely on a multi-task loss function based mainly on two terms. In particular, a depth term is in charge of minimize the discrepancy between the target image $I_t$ and an image $I_s$, warped as $\widetilde{I}_t^s$, from a monocular sequence while a semantic term is used to learn semantic labels from proxy label distilled by a pre-trained network.

**Depth term.** According to the self-supervised training paradigm proposed in [13], we adopt a photometric loss function consisting in a weighted combination between the Structural Dissimilarity Measure (DSSIM) and the standard $\mathcal{L}_1$ loss. In addition, a per-pixel minimum strategy [14] is used to solve occlusion/disocclusion by simply picking the minimum error between each pair $I_t$ and $I_s$ instead of averaging them. Thus, the photometric loss function is defined as:

$$\mathcal{L}_{ap}^D = \sum_p \min_s (\alpha \mathcal{L}_{DSSIM}(p) + (1-\alpha)||I_t(p) - \widetilde{I}_t^s(p)||_1) \quad (1)$$

where $p$ denotes pixel coordinates, $\widetilde{I}_t^s$ a source image $I_s$ warped according to estimated depth and pose and the $DSSIM$ loss function is computed as:

$$\mathcal{L}_{DSSIM}(p) = \frac{1 - SSIM(I_t(p), \widetilde{I}_t^s(p))}{2} \quad (2)$$

In our experiments, we set $\alpha = 0.85$.

---

[*]Joint first authorship.

1

| Layer | K | S | In/Out | Input |
|---|---|---|---|---|
| **Deep feature extractor (DSE)** | | | | |
| conv1a | 3 | 2 | 3/16 | input |
| conv1b | 3 | 1 | 16/16 | conv1a |
| conv2a | 3 | 2 | 16/32 | conv1b |
| conv2b | 3 | 1 | 32/32 | conv2a |
| conv3a | 3 | 2 | 32/64 | conv2b |
| conv3b | 3 | 1 | 64/64 | conv3a |
| conv4a | 3 | 2 | 64/128 | conv3b |
| conv4b | 3 | 1 | 128/128 | conv4a |
| conv5a | 3 | 2 | 128/256 | conv4b |
| conv5b | 3 | 1 | 256/256 | conv5a |
| **Estimator (E)** | | | | |
| conv1 | 3 | 1 | i_channels/64 | input |
| conv2 | 3 | 1 | 64/48 | conv1 |
| conv3 | 3 | 1 | 48/32 | conv2 |
| conv4 | 3 | 1 | 32/16 | conv3 |
| **Context (C)** | | | | |
| disp1 | 3 | 1 | i_channels/64 | input |
| disp2 | 3 | 1 | 64/32 | disp1 |
| disp3 | 3 | 1 | 32/16 | disp2 |
| disp | 3 | 1 | 16/1 | disp2 |
| **Disparity and Semantic Tower** | | | | |
| conv5 | 3 | 1 | i_channels/16 | E(conv5b) |
| disp5 | 3 | 1 | i_channels//1 | C(conv5) |
| conv4 | 3 | 1 | i_channels/16 | E(conv4b, disp5 ↑) |
| disp4 | 3 | 1 | i_channels//1 | C(conv4, conv5 ↑) + disp5 ↑ |
| conv3 | 3 | 1 | i_channels/16 | E(conv3b, disp4 ↑) |
| disp3 | 3 | 1 | i_channels//1 | C(conv3, conv4 ↑) + disp4 ↑ |
| conv2 | 3 | 1 | i_channels/16 | E(conv2b, disp3 ↑) |
| disp2 | 3 | 1 | i_channels//1 | C(conv2, conv3 ↑) + disp3 ↑ |
| conv1 | 3 | 1 | i_channels/16 | E(conv1b, disp2 ↑) |
| disp1 | 3 | 1 | i_channels//1 | C(conv1, conv2 ↑) + disp2 ↑ |
| sem | 3 | 1 | i_channels//1 | C(conv1, conv2 ↑) + disp2 ↑ |

| Layer | K | S | In/Out | Input |
|---|---|---|---|---|
| **Deep feature extractor (DFE)** | | | | |
| conv1a | 3 | 2 | 3/16 | input |
| conv1b | 3 | 1 | 16/16 | conv1a |
| conv2a | 3 | 2 | 16/32 | conv1b |
| conv2b | 3 | 1 | 32/32 | conv2a |
| conv3a | 3 | 2 | 32/64 | conv2b |
| conv3b | 3 | 1 | 64/64 | conv3a |
| conv4a | 3 | 2 | 64/128 | conv3b |
| conv4b | 3 | 1 | 128/128 | conv4a |
| **Pose Estimator** | | | | |
| conv1a | 3 | 1 | i_channels/128 | $DFE_t, DFE_s$ |
| conv1b | 3 | 2 | 128/128 | conv1a |
| conv2a | 3 | 1 | 128/256 | conv1b |
| conv2b | 3 | 2 | 256/256 | conv2a |
| pose | 1 | 1 | 256/6*N | conv2b |
| **Intrinsic Estimator** | | | | |
| focals | 1 | 1 | i_channels/2 | conv2b |
| offsets | 1 | 1 | i_channels/2 | conv2b |

Table 1. Detailed structure of the DSNet (left) and CamNet (right) modules in $\Omega$Net. The symbol "," means concatenation, while ↑ indicates upsampling.

A smoothness term is also used to penalize large disparity differences between adjacent pixels when the former do not co-occur with strong RGB gradients:

$$\mathcal{L}_{smooth} = \sum_p |\nabla D_t(p)| \cdot \left(e^{-|\nabla I_t(p)|}\right)^T \qquad (3)$$

Finally, we mask-out pixels whose appearance do not change between consecutive frames, which includes scenes with no relative motion. This has the effect of letting the network ignore pixels which move at the same velocity as the camera, and even to ignore whole frames when the camera stop moving. According to [14], this is accomplished by removing those pixels which have an unwarped photometric loss smaller than the corresponding warped photometric loss, *i.e.*

$$\mu = \min \mathcal{L}_{ap}^D(I_t, I_s) > \mathcal{L}_{ap}^D(I_t, I_t^s) \qquad (4)$$

**Semantic term.** The standard cross-entropy loss between the predicted and proxy pixel-wise semantic labels is used as semantic term:

$$\mathcal{L}_{sem} = -(S_t \log(S_P) + (1 - S_t) \log(1 - S_P)) \qquad (5)$$

where $S_t$ is the semantics predicted by DSNet and $S_P$ the ground-truth proxy label. Moreover, as proposed in [24] we employ a cross-task loss to tighten the link between the learning tasks dealing with depth and semantics:

$$\mathcal{L}_{cdd} = \sum_p sgn(|\nabla S_t(p)|) \cdot \left(e^{-|\nabla D_t(p)|}\right)^T \qquad (6)$$

Hence, the total loss used to train DSNet is a weighted combination of the above losses:

$$\mathcal{L} = \lambda_1 \mathcal{L}_{ap}^D + \lambda_2 \mathcal{L}_{smooth} + \lambda_3 \mathcal{L}_{sem} + \lambda_4 \mathcal{L}_{cdd} \tag{7}$$

where $\lambda_1$, $\lambda_2$, $\lambda_3$ and $\lambda_4$ are hyper-parameters. In our experiments, we set $\lambda_1 = 1$, $\lambda_2 = 0.1$, $\lambda_3 = 1$ and $\lambda_4 = 0.1$.

As described in the paper, for the Optical Flow we rely on a peculiar training schedule based on two components in $\Omega$Net, which are referred as OFNet and SD-OFNet.

**Optical Flow term.** We train a the first instance of the optical flow network (OFNet) using the same photometric loss as for DSNet:

$$\mathcal{L}_{ap}^{OF} = \sum_p \alpha \mathcal{L}_{DSSIM} + (1-\alpha) ||I_t - \widetilde{I}_t^s||_1 \tag{8}$$

In this case, however, $\widetilde{I}_t^s$ is warped according to estimated flow. Akin to DSNet, we set $\alpha = 0.85$.

**Self-Distilled Optical Flow term.** The self-distilled optical flow network (SD-OFNet), instead, is trained in a quite different manner. In fact, given the optical flow $F_{t \to s}$ predicted by OFNet, the rigid flow $F_{t \to s}^{rigid}$ and the mask $M$, we leverage on the optical flow in the regions where $F_{t \to s}$ and $F_{t \to s}^{rigid}$ are similar as well as on moving objects, while we rely on the rigid flow for the remaining areas (*e.g.*, occlusions due to camera motion). We can distinguish the former regions from the latter ones looking at $M$. Moreover, we also apply a photometric term $\phi$ on the predicted optical flow $SF_{t \to s}$. The final loss $\mathcal{L}$ to train SD-OFNet is given by:

$$\mathcal{L} = \sum \alpha_r \phi(SF_{t \to s}, F_{t \to s}^{rigid}) \cdot (1-M) + \alpha_d \phi(SF_{t \to s}, F_{t \to s}) \cdot M + \psi(I_t, \widetilde{I}_t^{SF}) \cdot M \tag{9}$$

During training, $F_{t \to s}$, $F_{t \to s}^{rigid}$, $M$ and the input images of SD-OFNet are randomly cropped to $416 \times 128$ before computing $\mathcal{L}$: in doing so, the errors at occluded areas in $F_{t \to s}$ due to camera motions, clearly visible in Figure 5, are less to appear and impact the training process. Finally, to ameliorate the photometric loss term, the image $\widetilde{I}^{SF}$ is obtained by padding the $SF_{t \to s}$ at first, which is predicted at $416 \times 128$, to original resolution (*e.g.*, $640 \times 192$), then using this flow to warp the full resolution $I_s$ at $I_t$ coordinates and finally extracting from this image the same crop as used before. This simple strategy allows to leverage on a complete image, since otherwise the cropped image would suffer from motion occlusions near boundaries. Moreover, we highlight that SD-OFNet is initialized to the OFNet weights, *i.e.* those found during the above described OFNet training based on the standard photometric loss, and then, when training SD-OFNet, only its weights are updated, *i.e.* OFNet is kept frozen.

## 3. Monocular Depth Estimation

In this section, we provide more insights on $\Omega$Net performance concerning depth estimation, in particular by reporting comparison with state-of-the-art methods trained with *stronger* supervision, a more detailed analysis about the errors computed at different depth ranges and a reproducibility study about DSNet.

### 3.1. Comparison with more methods on the KITTI Eigen split

In this section, we report additional comparisons on the Eigen's KITTI test split [11]. In particular, we compare $\Omega$Net to state-of-the-art frameworks trained with stronger forms of supervision, *i.e.* stereo pairs, stereo videos or proxy labels. Differently from these approaches, we do not apply any post-processing step to further improve predictions. As highlighted in Table 3, we can notice how our method is comparable and, in most cases performs better, wrt other self-supervised depth-from-mono architectures trained on stereo pairs/stereo videos. Moreover, we point out that we outperform frameworks running online adaptation on the testing set [3, 7] on most metrics. Only semi-supervised methods at the bottom of the table [32, 27, 30] are in general more effective, because of the much stronger supervision from traditional stereo algorithms deployed during training.

### 3.2. Error at different depth ranges

In Table 3, we report more data supporting the claim that DSNet produces more accurate depth estimates at long distances with respect to other strategies such as [14] or even replacing our architecture with a much more complex one [34] based on a ResNet-50 backbone. We deeply looked into this and ascribe this finding to more complex models producing *oversmoothed* depth maps. In particular, in our experiments, we noticed that our shallow network tends to produce much sharper estimates compared to models having many more parameters. *Over-smoothing* produces better qualitative predictions and higher accuracy at short ranges, but it degrades depth accuracy at long distances, as we can observe in the table.

| | | | | | | | | Lower is better | | | | Higher is better | | |
|---|---|---|---|---|---|---|---|---|---|---|---|---|---|---|
| Method | M | S | V | P | A | I | CS | Abs Rel | Sq Rel | RMSE | RMSE log | $\delta < 1.25$ | $\delta < 1.25^2$ | $\delta < 1.25^3$ |
| Zhou et al. [37] | | | ✓ | | | | ✓ | 0.198 | 1.836 | 6.565 | 0.275 | 0.718 | 0.901 | 0.960 |
| Godard et al. [14] | | | ✓ | | | ✓ | | 0.115 | 0.903 | 4.863 | 0.193 | 0.877 | 0.959 | 0.981 |
| Godard et al. [14] | | | ✓ | | | | | 0.132 | 1.044 | 5.142 | 0.210 | 0.845 | 0.948 | 0.977 |
| Godard et al. [14] (1024 × 320) | | | ✓ | | | ✓ | | **0.115** | 0.882 | 4.701 | 0.190 | **0.879** | **0.961** | 0.982 |
| Zhou et al. [36] | | | ✓ | | | ✓ | | 0.121 | 0.837 | 4.945 | 0.197 | 0.853 | 0.955 | 0.982 |
| Mahjourian et al. [19] | | | ✓ | | | | ✓ | 0.159 | 1.231 | 5.912 | 0.243 | 0.784 | 0.923 | 0.970 |
| Yang et al. [33] | | | ✓ | | | | ✓ | 0.159 | 1.345 | 6.254 | 0.247 | - | - | - |
| Wang et al. [28] | | | ✓ | | | | ✓ | 0.151 | 1.257 | 5.583 | 0.228 | 0.810 | 0.936 | 0.974 |
| Bian et al. [2] | | | ✓ | | | | | 0.128 | 1.047 | 5.234 | 0.208 | 0.846 | 0.947 | 0.970 |
| Yin et al. [34] | ✓ | | ✓ | | | | ✓ | 0.153 | 1.328 | 5.737 | 0.232 | 0.802 | 0.934 | 0.972 |
| Zou et al. [38] | ✓ | | ✓ | | | | ✓ | 0.146 | 1.182 | 5.215 | 0.213 | 0.818 | 0.943 | 0.978 |
| Chen et al. [7] | ✓ | | ✓ | | | | | 0.135 | 1.070 | 5.230 | 0.210 | 0.841 | 0.948 | 0.980 |
| Luo et al. [18] | ✓ | | ✓ | | | | | 0.141 | 1.029 | 5.350 | 0.216 | 0.816 | 0.941 | 0.976 |
| Ranjan et al. [1] | ✓ | | ✓ | | | | ✓ | 0.139 | 1.032 | 5.199 | 0.213 | 0.827 | 0.943 | 0.977 |
| Xu et al. [31] | | | ✓ | ✓ | | | | 0.138 | 1.016 | 5.352 | 0.217 | 0.823 | 0.943 | 0.976 |
| Casser et al. [3] | | | ✓ | ✓ | | | | 0.141 | 1.026 | 5.290 | 0.215 | 0.816 | 0.945 | 0.979 |
| Gordon et al. [15] | ✓ | | ✓ | ✓ | | | | 0.128 | 0.959 | 5.230 | - | - | - | - |
| ΩNet(416 × 128) | | | ✓ | ✓ | | | | 0.134 | 0.893 | 5.137 | 0.208 | 0.829 | 0.946 | 0.979 |
| ΩNet(640 × 192) | | | ✓ | ✓ | | | | 0.126 | 0.835 | 4.937 | 0.199 | 0.844 | 0.953 | 0.982 |
| ΩNet(1024 × 320) | | | ✓ | ✓ | | | | 0.125 | 0.805 | 4.795 | 0.195 | 0.849 | 0.955 | 0.983 |
| ΩNet(416 × 128) | | | ✓ | ✓ | | ✓ | | 0.126 | 0.862 | 4.963 | 0.199 | 0.846 | 0.952 | 0.981 |
| ΩNet(640 × 192) | | | ✓ | ✓ | | ✓ | | 0.120 | 0.792 | 4.750 | 0.191 | 0.856 | 0.958 | 0.984 |
| ΩNet(1024 × 320) | | | ✓ | ✓ | | ✓ | | 0.118 | **0.748** | 4.608 | **0.186** | 0.865 | **0.961** | 0.985 |
| ΩNet(768 × 384) † | | | ✓ | ✓ | | ✓ | | 0.184 | 1.565 | 6.456 | 0.243 | 0.742 | 0.920 | 0.974 |
| Casser et al. [3] (+ Online Ref.) | | | ✓ | ✓ | | | | 0.109 | 0.825 | 4.750 | 0.187 | 0.874 | **0.958** | 0.983 |
| Chen et al. [7] (+ Online Ref.) | | | ✓ | | | | | **0.099** | 0.796 | 4.743 | 0.186 | **0.884** | 0.955 | 0.979 |
| Poggi et al. [22] | | ✓ | | | | | ✓ | 0.146 | 1.291 | 5.907 | 0.245 | 0.801 | 0.926 | 0.967 |
| Poggi et al. [23] | | ✓ | | | | | ✓ | 0.111 | 0.849 | 4.822 | 0.202 | 0.865 | 0.952 | 0.978 |
| Pillai et al. [21] | | ✓ | | | | | | 0.112 | 0.875 | 4.958 | 0.207 | 0.852 | 0.947 | 0.977 |
| Godard et al. [14] | | ✓ | | | | ✓ | | **0.106** | 0.806 | 4.630 | **0.193** | **0.876** | 0.958 | **0.980** |
| Godard et al. [13] | | ✓ | | | | | ✓ | 0.114 | 0.898 | 4.935 | 0.206 | 0.861 | 0.949 | 0.976 |
| Zhang et al. [35] | | ✓ | ✓ | | | | | 0.135 | 1.132 | 5.585 | 0.229 | 0.820 | 0.933 | 0.971 |
| Luo et al. [18] | ✓ | ✓ | | | | | | 0.127 | 0.936 | 5.008 | 0.209 | 0.841 | 0.946 | 0.979 |
| Yang et al. [32] | | ✓ | ✓ | ✓ | | | | 0.097 | 0.734 | 4.442 | 0.187 | 0.888 | 0.958 | 0.980 |
| Watson et al. [30] | | ✓ | | ✓ | | ✓ | | **0.096** | 0.710 | 4.393 | 0.185 | **0.890** | **0.962** | 0.981 |
| Tosi et al. [27] | | ✓ | | ✓ | | | | **0.096** | 0.673 | 4.351 | **0.184** | **0.890** | 0.961 | 0.981 |

Table 2. Quantitative evaluation on the Eigen test set of the KITTI dataset [12] for self-supervised monocular depth estimation methodologies. S: stereo pairs, V: video sequence, P: depth proxy labels, A: additional information, I: feature extractors pre-trained on ImageNet [9] or CS: Cityscapes [8]. †Trained on CS and tested on KITTI without any fine-tuning.

4

| Method | Cap (m) | Abs Rel | Sq Rel | RMSE | RMSE log |
|---|---|---|---|---|---|
| Godard et al. [14] | 0-8 | **0.059** | **0.062** | 0.503 | **0.082** |
| **Ours**† | 0-8 | 0.060 | 0.063 | **0.502** | **0.082** |
| **Ours** | 0-8 | 0.062 | 0.065 | 0.517 | 0.085 |
| Godard et al. [14] | 0-15 | **0.083** | 0.173 | 1.178 | 0.125 |
| **Ours**† | 0-15 | **0.083** | **0.168** | **1.148** | **0.122** |
| **Ours** | 0-15 | 0.084 | 0.169 | 1.156 | 0.124 |
| Godard et al. [14] | 0-30 | **0.111** | 0.470 | 2.561 | 0.172 |
| **Ours**† | 0-30 | **0.111** | 0.442 | 2.513 | **0.169** |
| **Ours** | 0-30 | **0.111** | 0.425 | **2.463** | **0.169** |
| Godard et al. [14] | 0-50 | **0.125** | 0.788 | 3.946 | 0.198 |
| **Ours**† | 0-50 | 0.127 | 0.762 | 4.020 | 0.199 |
| **Ours** | 0-50 | **0.124** | **0.702** | **3.836** | **0.195** |
| Godard et al. [14] | 0-80 | 0.132 | 1.044 | 5.142 | 0.210 |
| **Ours**† | 0-80 | 0.134 | 1.074 | 5.451 | 0.213 |
| **Ours** | 0-80 | **0.126** | **0.835** | **4.937** | **0.199** |

Table 3. Depth errors at different depth ranges. †indicates that our depth network has been replaced with the heavy-weight [34] backbone based on the ResNet50 architecture.

### 3.3. Reproducibility

We perform three independent training of our architecture to assess upon its reproducibility. Table 4 shows how our architecture produces the same results with negligible variance due to the randomness factors in training, *i.e.* initialization, data shuffle and augmentation.

| Resolution | Lower is better | | | | Higher is better | | |
|---|---|---|---|---|---|---|---|
| | Abs Rel | Sq Rel | RMSE | RMSE log | $\delta < 1.25$ | $\delta < 1.25^2$ | $\delta < 1.25^3$ |
| $640 \times 192$ | 0.120 | 0.792 | 4.750 | 0.191 | 0.856 | 0.958 | 0.984 |
| $640 \times 192$ | 0.122 | 0.799 | 4.749 | 0.191 | 0.856 | 0.958 | 0.984 |
| $640 \times 192$ | 0.121 | 0.795 | 4.755 | 0.192 | 0.855 | 0.957 | 0.983 |

Table 4. Three independent runs of our $\Omega$Net(DSNet) result in slightly different models on the KITTI Eigen split.

## 4. Semantic Segmentation

In this section we report more detailed semantic segmentation results. Purposely, we use the following metrics:

1. **IoU**: Intersection over Union for pixel-wise segmentation calculated for each class or category, as defined in [8].

2. **mIoU$_{class}$**: mean IoU for the the 19 training classes used in CityScapes [8]: road, sidewalk, building, wall, fence, pole, traffic light, traffic sign, vegetation, terrain, sky, person, rider, car, truck, bus, train, motorcycle and bicycle.

3. **mIoU$_{category}$**, mean IoU considering the 7 macro-classes defined in CityScapes [8]: flat, construction, object, nature, sky, human, vehicles.

4. **Pixel Accuracy (Acc.)**: ratio between the correct and the total pixel predictions without considering any specific class or category.

### 4.1. Generalization across Datasets

In Tables 5 and 6 we validate with thorough experiments the motivation behind our better generalization across datasets compared to other state-of-the-art methods for real-time semantic segmentation. In this study we train on CityScapes and test on KITTI, reporting in Table 5 the IoU for the 19 classes, the mIoU$_{class}$ and the pixel Acc. In Table 6 we report the IoU for the 7 categories and the mIoU$_{category}$. We refer with CS(S) methods trained on 2975 CityScapes images and with CS(P) methods trained on 22,973 proxy labels produced by [6]. To evaluate the performance of [16, 4] we used the official code and pre-trained weights available online. Our DSNet differs from other methods by three factors: 1) the architecture, 2) the training protocol exploiting proxy labels instead of ground truths and 3) the joint reasoning about geometry and semantics.

Regarding the tests on KITTI, our architecture trained only for semantic segmentation, namely Semantic Network or SNet, achieves good performance in Acc. but turns out worse than [4] for other metrics. On the other hand, it is worth to notice that training SNet with CS(P) allows our method to achieve a great performance boost in all metrics with respect to CS(P) (rows 8 vs 9). Finally, we can notice how DSNet achieves results comparable to SNet. This confirms the findings in [24], that joint reasoning about depth and semantics is more beneficial to the former task.

5

| Method | Train | Test | road | sidewalk | building | wall | fence | pole | traffic light | traffic sign | vegetation | terrain | sky | person | rider | car | truck | bus | train | motorcycle | bicycle | mIoU$_{class}$ | Acc. |
|---|---|---|---|---|---|---|---|---|---|---|---|---|---|---|---|---|---|---|---|---|---|---|---|
| DABNet [16] | CS(S) | CS | 97.05 | 82.86 | 91.01 | 48.20 | 55.56 | 59.30 | 63.12 | 72.76 | 91.58 | **61.24** | 93.50 | 77.96 | 54.70 | 93.28 | 53.06 | 71.01 | 27.77 | 56.00 | 72.91 | 69.62 | 94.62 |
| FCHardNet [4] | CS(S) | CS | **97.39** | **84.40** | **92.31** | **53.83** | **62.90** | **64.28** | **68.21** | **78.06** | **91.85** | 59.82 | **94.91** | **80.81** | **60.55** | **94.85** | **72.70** | **82.15** | **76.45** | **59.97** | **75.49** | **76.37** | **95.35** |
| ΩNet(SNet) | CS(S) | CS | 93.69 | 65.66 | 83.46 | 23.57 | 20.57 | 40.11 | 35.32 | 47.77 | 86.62 | 44.22 | 89.94 | 56.00 | 23.00 | 84.98 | 17.22 | 1.22 | 0.00 | 17.11 | 52.82 | 46.49 | 89.56 |
| ΩNet(SNet) | CS(P) | CS | 95.97 | 77.23 | 87.96 | 38.37 | 42.62 | 47.82 | 48.15 | 60.44 | 89.73 | 54.97 | 92.62 | 65.87 | 36.96 | 90.57 | 25.19 | 0.06 | 0.00 | 25.53 | 61.06 | 54.80 | 92.45 |
| ΩNet(DSNet) | CS(P) | CS | 96.00 | 77.46 | 88.30 | 41.84 | 41.68 | 48.74 | 47.80 | 59.24 | 89.61 | 53.89 | 92.57 | 66.29 | 38.61 | 90.61 | 27.39 | 0.37 | 0.00 | 18.01 | 62.78 | 54.80 | 92.50 |
| DABNet [16] | CS(S) | K | 79.02 | 19.07 | 58.38 | 18.04 | **30.73** | 40.61 | 44.24 | 41.67 | 80.87 | 48.76 | 76.61 | 13.39 | 0.17 | 63.30 | 21.32 | 8.21 | 19.81 | 1.29 | 7.04 | 35.40 | 80.50 |
| FCHardNet [4] | CS(S) | K | 75.66 | 32.65 | **78.51** | 13.16 | 28.46 | 51.33 | **57.16** | 55.58 | 81.06 | 45.59 | 91.43 | 23.84 | 12.19 | 58.86 | **24.91** | **34.89** | **68.71** | 4.66 | 11.38 | **44.74** | 72.07 |
| ΩNet(SNet) | CS(S) | K | 83.31 | 33.39 | 66.57 | 12.15 | 20.18 | 44.20 | 37.76 | 32.35 | 84.46 | 58.79 | 88.70 | 24.66 | 13.55 | 76.09 | 12.62 | 2.09 | 0.10 | 1.15 | 12.64 | 37.09 | 84.94 |
| **ΩNet(SNet)** | CS(P) | K | **88.73** | **47.85** | 77.01 | **19.72** | 30.65 | 47.34 | 53.63 | 43.16 | 86.65 | **67.97** | 94.49 | 24.81 | 29.39 | 80.68 | 14.88 | 0.53 | 0.00 | 3.05 | 12.30 | 43.31 | 88.76 |
| **ΩNet(DSNet)** | CS(P) | K | 87.89 | 46.64 | 77.48 | 18.55 | 29.65 | **48.73** | 51.12 | 40.52 | **86.66** | 63.54 | **95.06** | 29.79 | **34.74** | **82.03** | 12.77 | 0.63 | 0.00 | **7.60** | **18.82** | 43.80 | **88.31** |

Table 5. IoU on 19 training classes, mIoU$_{class}$ and pixel accuracy (Acc.) results of ΩNet against state of the art method training on CS and tested on CS or K. Better generalization from CS to K thanks to our proxy labels training protocol.

| Method | Train | Test | flat | construction | object | nature | sky | human | vehicle | mIoU$_{category}$ |
|---|---|---|---|---|---|---|---|---|---|---|
| DABNet [16] | CS(S) | CS | 97.93 | 91.69 | 65.90 | 92.03 | 93.50 | 79.59 | 92.25 | 87.56 |
| FCHardNet [4] | CS(S) | CS | **98.19** | **92.55** | **70.77** | **92.27** | **94.91** | **82.31** | **93.54** | **89.22** |
| ΩNet(SNet) | CS(S) | CS | 96.34 | 84.29 | 44.37 | 86.85 | 89.94 | 60.13 | 83.77 | 77.96 |
| ΩNet(SNet) | CS(P) | CS | 97.40 | 88.80 | 53.80 | 90.19 | 92.62 | 69.08 | 88.47 | 82.88 |
| ΩNet(DSNet) | CS(P) | CS | 97.38 | 88.76 | 53.91 | 89.93 | 92.57 | 69.27 | 88.61 | 82.92 |
| DABNet [16] | CS(S) | K | 83.41 | 59.07 | 46.41 | 84.30 | 76.61 | 17.05 | 63.61 | 61.49 |
| FCHardNet [4] | CS(S) | K | 80.89 | 75.35 | 58.68 | 88.11 | 91.43 | 24.62 | 58.33 | 68.20 |
| **ΩNet(SNet)** | CS(S) | K | 87.93 | 63.92 | 45.79 | 85.47 | 88.70 | 31.02 | 69.95 | 67.54 |
| **ΩNet(SNet)** | CS(P) | K | **91.97** | **74.95** | 52.29 | 89.80 | 94.49 | 29.28 | **81.83** | 73.52 |
| **ΩNet(DSNet)** | CS(P) | K | 91.42 | 74.84 | **53.35** | 89.36 | **95.06** | **35.45** | 80.69 | **74.31** |

Table 6. IoU on 7 training categories and, mIoU$_{category}$ results of ΩNet against state of the art method training on CS and tested on CS or K. Better generalization from CS to K thanks to our proxy labels training protocol.

### 4.2. Proxy Semantic Network

We evaluate the performance of the proxy semantic network. We employ DPC [6], pre-trained on CityScapes with the 2975 training ground truths. We report in Table 7 the testing results on the 500 and 200 images belonging to CityScapes validation set and the KITTI training datasets, respectively. Even though DPC [6] achieves impressive performance both on CityScapes as well as in generalizing to KITTI, it is a huge network unable to run in real-time (*i.e.*, it approximately delivers 3.5 fps on $768 \times 384$ images).

| Method | Train | Test | mIoU$_{class}$ | mIoU$_{category}$ | Acc. |
|---|---|---|---|---|---|
| DPC[5] - Proxy | CS(S) | CS | 80.22 | 90.73 | 95.99 |
| DPC[5] - Proxy | CS(S) | K | 58.75 | 81.30 | 90.21 |

Table 7. Semantic segmentation performances of the proxy semantic network [6] on CS and K datasets.

### 4.3. Priors Evaluation on KITTI

When we produce the priors used during training and, at prediction time, to create the $M_t^{mot}$, we split the 19 classes in static and potentially dynamic ones according to the following scheme:

1. **Static**: road, sidewalk, building, wall, fence, pole, traffic light, traffic sign, vegetation, terrain, sky

2. **Potentially dynamic**: person, rider, car, truck, bus, train

As among our objectives is to obtain a good motion segmentation mask, we are interested in evaluating the quality of our semantic segmentation predictions in terms of how they are amenable to producing good estimated priors according to the mapping defined above. We evaluate our DSNet trained on CityScapes+KITTI in the 200 KITTI images which provides semantic labels. We obtain a pixel accuracy of 98.50% while a 98.40% IoU for the static classes and a 80.99% for the potentially dynamic classes for a global 89.64% mIoU. It is worth noticing that, even though our segmentation is not able to perform a precise class segmentation, it yields excellent binary priors that turns out key to performance for motion segmentation.

# 5. Optical Flow Estimation

## 5.1. Comparison with more methods on the KITTI 2015 split

In Table 8 we include additional results from our main competitors to allow for a more comprehensive analysis. In particular, we report additional experiments from [1], in which the authors exploit a different combination of depth and optical flow networks, and from [29], that demonstrate how using stereo pairs at training time allows to obtain much better results on rigid regions. Nonetheless, it can be noticed that our network still outperforms existing monocular multi-task strategies by a large margin.

|  |  | train |  |  | test |
|---|---|---|---|---|---|
| Method | Dataset | Noc | All | F1 | F1 |
| Meister*et al*. [20] - C | SYN + K | - | 8.80 | 28.94% | 29.46% |
| Meister *et al*. [20] - CSS | SYN + K | - | 8.10 | 23.27% | 23.30% |
| Zou *et al*. [38] | SYN + K | - | 8.98 | 26.0% | 25.70% |
| Ranjan *et al*. [1] - DispResNet + PWC | SYN + K | - | 5.66 | 20.93% | 25.27% |
| Wang *et al*. [29] (Ego-motion) ** | K | - | 10.69 | - | 32.34% |
| Wang *et al*. [29] (Full) ** | K | - | 5.58 | - | 18.00% |
| Ren *et al*. [25] | K | - | 16.79 | 36.00% | 39.00% |
| Yin *et al*. [34] | K | 8.05 | 10.81 | - | - |
| Chen *et al*. [7] † | K | 5.40 | 8.95 | - | - |
| Chen *et al*. [7] (online) † | K | 4.86 | 8.35 | - | - |
| Ranjan *et al*. [1] - DispNet + FlowNetC | K | - | 7.76 | - | - |
| Ranjan *et al*. [1] - DispResNet + PWC | K | - | 6.21 | 26.41% | - |
| Luo *et al*. [18] | K | 5.84 | - | 21.56% | - |
| Luo *et al*. [18] * | K | 5.43 | - | 20.61% | - |
| $\Omega$**Net** (Ego-motion) | K | 11.72 | 13.50 | 51.22% | - |
| $\Omega$**Net**(OFNet) | K | 3.48 | 11.61 | 25.78% | - |
| $\Omega$**Net**(SD-OFNet) | K | **3.29** | **5.39** | **20.00**% | **19.47%** |

Table 8. We report percentage of erroneous pixels (F1 score) and average end-point error over all pixels (All) and non-occluded pixels (Noc) on the KITTI 2015 flow dataset. We indicate with †feature extractors pre-trained on ImageNet, SYN as the SYNTHIA [26] dataset, CS for the Cityscapes dataset, multi-task methods *trained on stereo pair and ** using stereo at testing time.

# 6. Pose Estimation

We validate the performance of our framework on pose estimation on the KITTI odometry split, which provides ground-truth camera poses obtained with IMU/GPS readings for 11 driving sequences, indexed from 00 to 08 for training and 09-10 for testing purposes. As in [14], we have not changed our architecture for this specific task but simply trained it from scratch on new training sequences without known intrinsic parameters. We compare our model with learned camera intrinsic parameters with several monocular self-supervised methods on the two sequences of KITTI odometry test split. All of the results, summarized in 9, are evaluated by optimizing the scaling factor to align with the ground-truth to address the inherent scale ambiguity.

| Method | Frames | Sequence 09 | Sequence 10 |
|---|---|---|---|
| Zhou *et al*. [37] | 5 | $0.021 \pm 0.017$ | $0.020 \pm 0.015$ |
| Ranjan *et al*. [1] | 5 | $0.012 \pm 0.007$ | $0.012 \pm 0.008$ |
| Yin *et al*. [34] | 5 | $0.012 \pm 0.007$ | $0.012 \pm 0.009$ |
| ORB-Slam | 3 | $0.014 \pm 0.008$ | $0.012 \pm 0.011$ |
| Casser *et al*. [3] | 3 | $0.011 \pm 0.006$ | $0.011 \pm 0.010$ |
| Zou *et al*. [3] | 3 | $0.017 \pm 0.008$ | $0.015 \pm 0.009$ |
| Luo *et al*. [18] | 3 | $0.013 \pm 0.007$ | $0.012 \pm 0.008$ |
| Godard *et al*. [14] | 2 | $0.017 \pm 0.008$ | $0.015 \pm 0.010$ |
| **Ours** † | 2 | $0.020 \pm 0.013$ | $0.017 \pm 0.011$ |

Table 9. Absolute Trajectory Error (ATE) of pose estimation evaluated on the KITTI odometry split sequences 09-10. †indicates strategies trained with unkwnown camera intrinsics.

# 7. Motion Segmentation

## 7.1. Threshold analysis

In Figure 1, we present an ablation study dealing with the motion segmentation task. In the main paper, to be consistent with other methodologies, we set the threshold $\tau$ used for the evaluation to $0.5$. However, we point out that a careful tuning of such threshold can improve the overall motion segmentation accuracy. In particular, we can notice how the best configuration for our predictions is obtained using a larger threshold. Indeed, we found out that the best trade-off between the mean accuracy and the mean IoU is achieved by setting the threshold value to $0.7$ (in this case the Mean Acc is $0.91$ while Mean IoU is $0.77$).

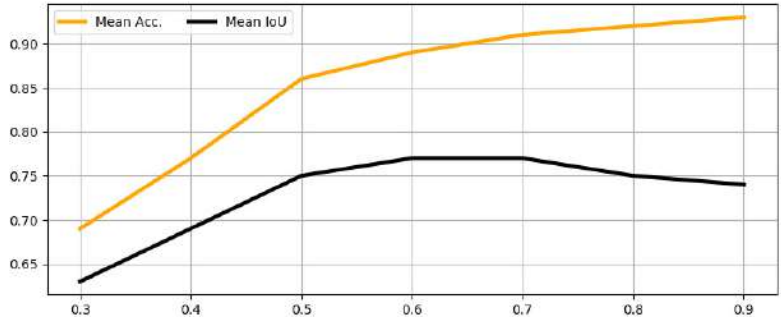

Figure 1. Mean Acc. and mIoU varying the threshold used to compute the motion segmentation $M_t^{mot}$.

## 7.2. Evaluation for KITTI only on Cars

We conduct an additional study to evaluate our motion segmentation masks only on pixels belonging to Cars, as proposed in [1]. In Table 10 we evaluate the IoU for static and dynamic cars yielded by $\Omega$Net and [1] on the 200 KITTI images endowed with ground truth for the motion segmentation task. We notice that our $M_t^{mot}$ outperforms [1] in all metrics (rows 1 vs 2 and 3) for all thresholds. Moreover, we point out that in this test configuration the contribution given to the motion segmentation by our estimated semantics is almost negligible as car regions are already extracted by using KITTI ground truths. Therefore, we test also our motion probability $P_t$ alone, showing that it is superior to [1] even without the help provided by the estimated semantics.

| Method | Threshold | Overall | Static Cars | Moving Cars |
|---|---|---|---|---|
| Ranjan [1] | - | 56.94 | 55.77 | 58.11 |
| $\Omega$Net $M_t^{mot}$ | 0.5 | **63.98** | **64.16** | 63.79 |
| $\Omega$Net $M_t^{mot}$ | 0.7 | 63.97 | 64.15 | 63.79 |
| $\Omega$Net $P_t$ | 0.5 | 63.67 | 62.58 | 64.77 |
| $\Omega$Net $P_t$ | 0.7 | 62.66 | 58.42 | **66.89** |

Table 10. Motion Segmentation Results. IoU scores on KITTI 2015 training dataset images computed only over car pixels.

# 8. Runtime

In this section we report additional runtime results on the three different devices used in the main paper, that is: an NVIDIA Titan Xp GPU, an Intel i7-7700K CPU and an NVIDIA Jetson TX2 GPU. In Table 11, we show further timings by varying the input image resolution of our architecture. It can be noticed how $\Omega$Net achieves real-time results (*i.e.* 27.9) on the Titan Xp GPU even with the largest image size $1024 \times 320$, reaching about 2 FPS on the Jetson Tx2 embedded device with the same input configuration.

|           |     | \multicolumn{5}{c}{$416 \times 128$} | \multicolumn{5}{c}{$640 \times 192$} | \multicolumn{5}{c}{$1024 \times 320$} |
|           | W   | D     | DS    | OF    | Cam   | O    | D     | DS    | OF   | Cam   | O    | D    | DS   | OF   | Cam   | O    |
|-----------|-----|-------|-------|-------|-------|------|-------|-------|------|-------|------|------|------|------|-------|------|
| Jetson TX2 | 15  | 20.2  | 17.9  | 8.9   | 54.1  | 7.1  | 12.5  | 10.3  | 6.5  | 49.2  | 4.5  | 6.4  | 5.3  | 3.2  | 26.31 | 2.0  |
| i7-7700K  | 91  | 10.9  | 9.1   | 11.0  | 60.1  | 5.5  | 5.0   | 4.2   | 4.9  | 31.4  | 2.4  | 1.9  | 1.6  | 1.8  | 13.2  | 0.9  |
| Titan XP  | 250 | 250.7 | 212.4 | 152.6 | 550.7 | 90.5 | 170.2 | 134.1 | 94.1 | 446.7 | 57.4 | 86.0 | 64.5 | 44.5 | 251.0 | 27.9 |

Table 11. Runtime analysis on different harware devices. For each device we report the power consumption in Watt and the FPS by varying input resolution. D: Depth, S: Semantic, OF: Optical Flow, Cam: camera pose, O: Overall architecture.

## 9. Qualitative results

In Figures 2, 3, 4, 5, 6, 7, 8, we provide qualitative results of our architecture on the standard datasets used in the main paper, such as KITTI and CityScapes. We refer the reader to the captions for description and comments related to each example.

### 9.1. Results on a YouTube Video

Furthermore, to prove that our network can be trained on unconstrained monocular sequences with unknown camera parameters and without semantic ground-truth labels, we downloaded from YouTube an online video captured by a moving camera consisting of 130K images depicting an urban scenario. Then, we generated proxy semantic labels using [6] and trained $\Omega$Net(DSNet) to learn depth, pose, semantics and camera intrinsics. Figure 9, show qualitative results yielded by $\Omega$Net on this unconstrained monocular video.

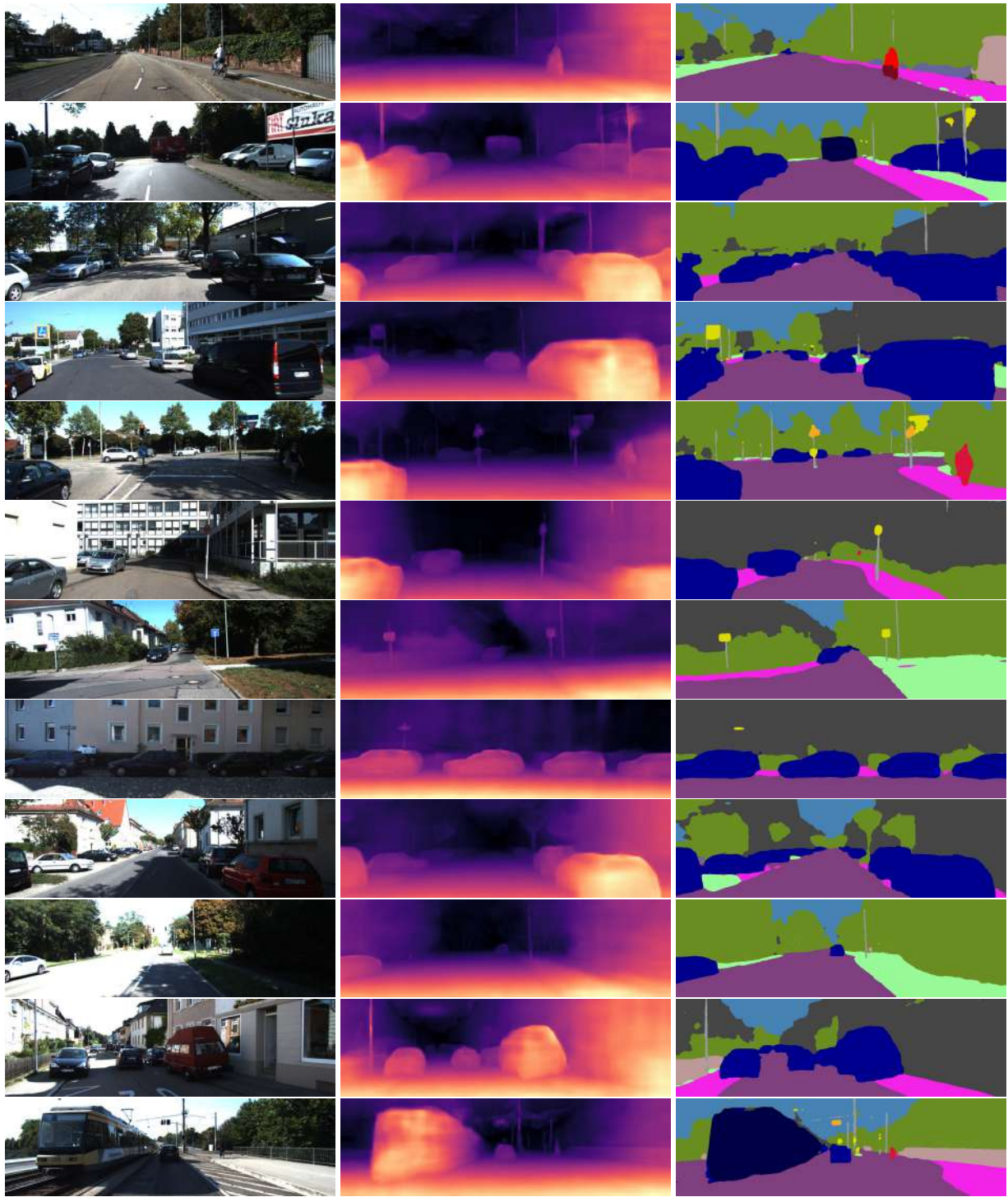

Figure 2. Qualitative results of our ΩNet(DSNet) on the KITTI Eigen split. From left to right we show image, the predicted single-image depth map and the predicted semantic segmentation of the scene.

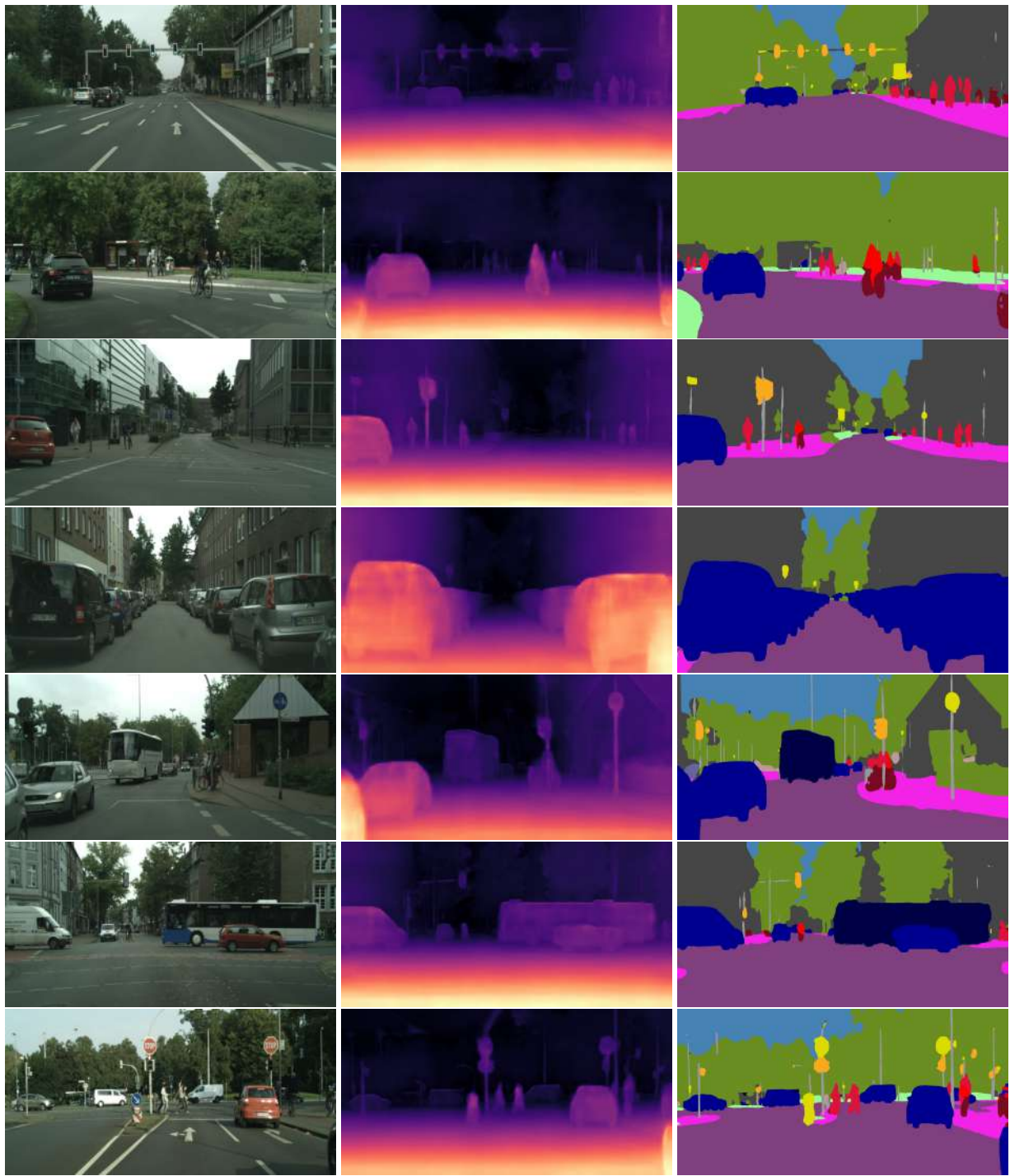

Figure 3. Qualitative results of our network ΩNet(DSNet) on the Cityscapes dataset. From left to right, the input image, single-view depth and the semantic prediction of our model.

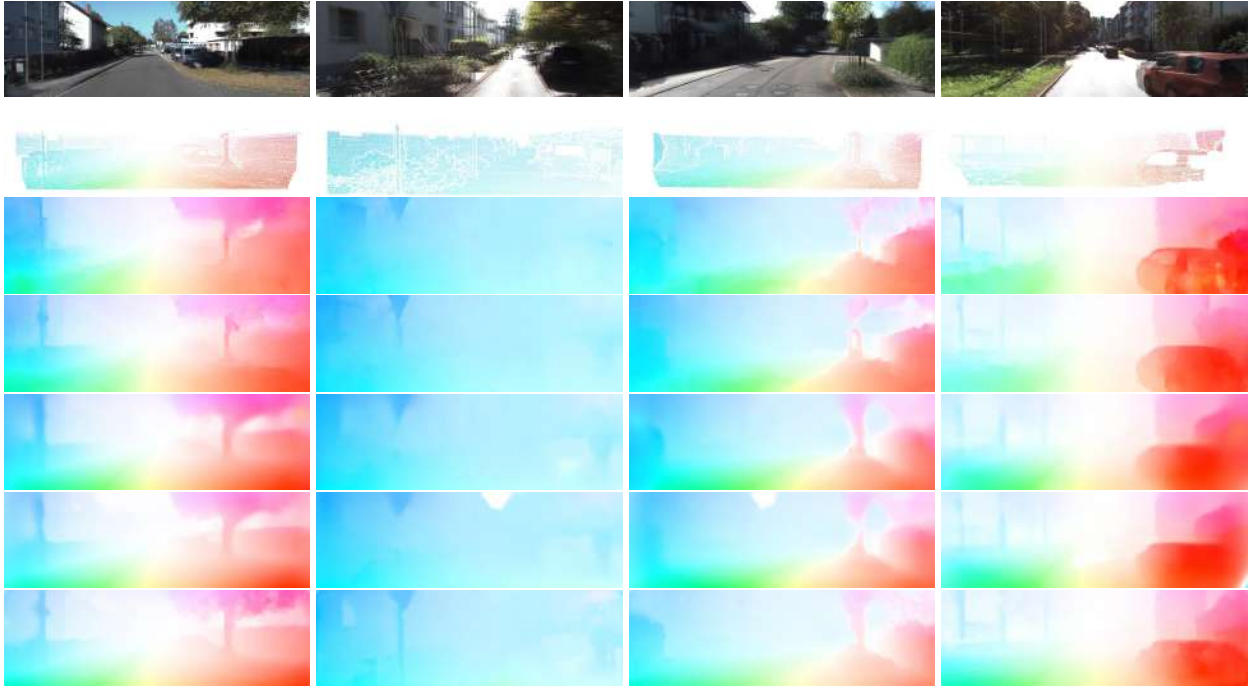

Figure 4. Qualitative results on optical flow estimation using the KITTI 2015 dataset. From top to bottom we show image, ground-truth labels, FlowNetS [10], FlowNetC [10], Unflow [20], DF-Net [38] and our ΩNet(SD-OFNet).

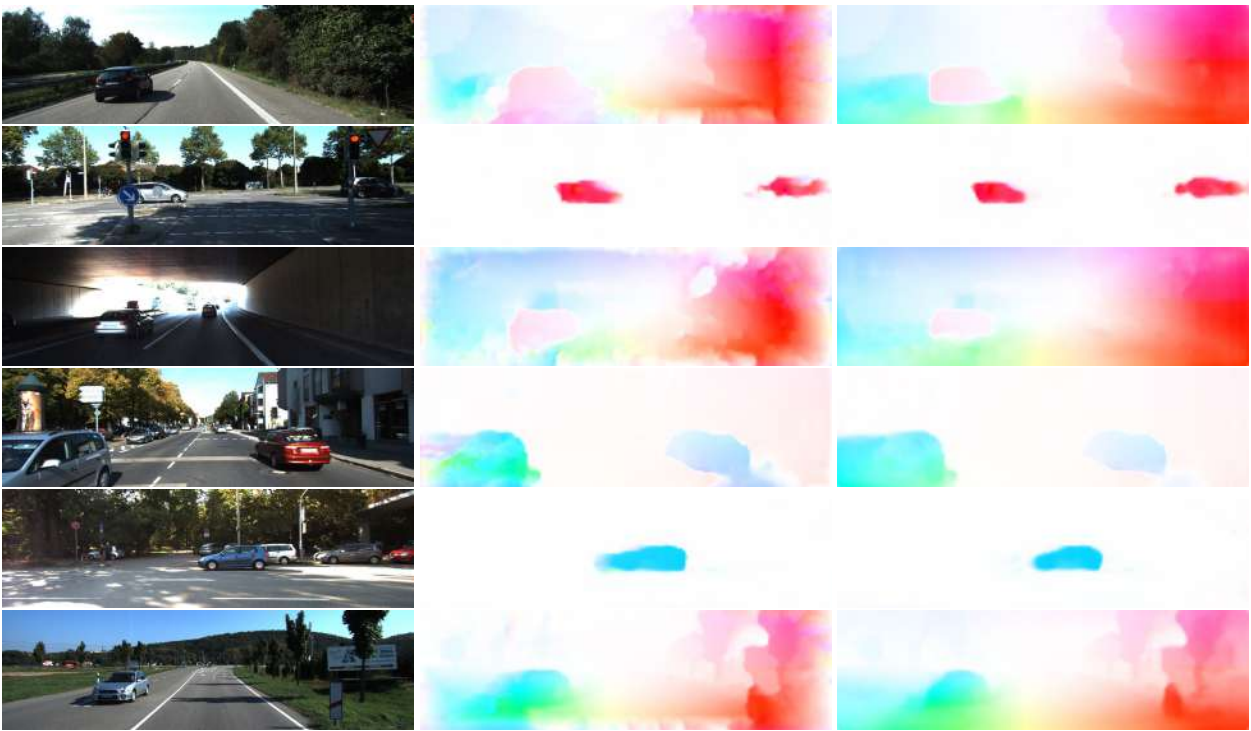

Figure 5. Qualitative comparison between the initial optical flow network ΩNet(OFNet) and the self-distilled one ΩNet(SD-OFNet) obtained with our strategy on the KITTI 2015 dataset. From left to right we show image, ΩNet(OFNet) and ΩNet(SD-OFNet) results respectively. It can be noticed that the proposed self-distilled paradigm greatly alleviates motion boundaries occlusions and improves details in the final optical flow predictions.

12

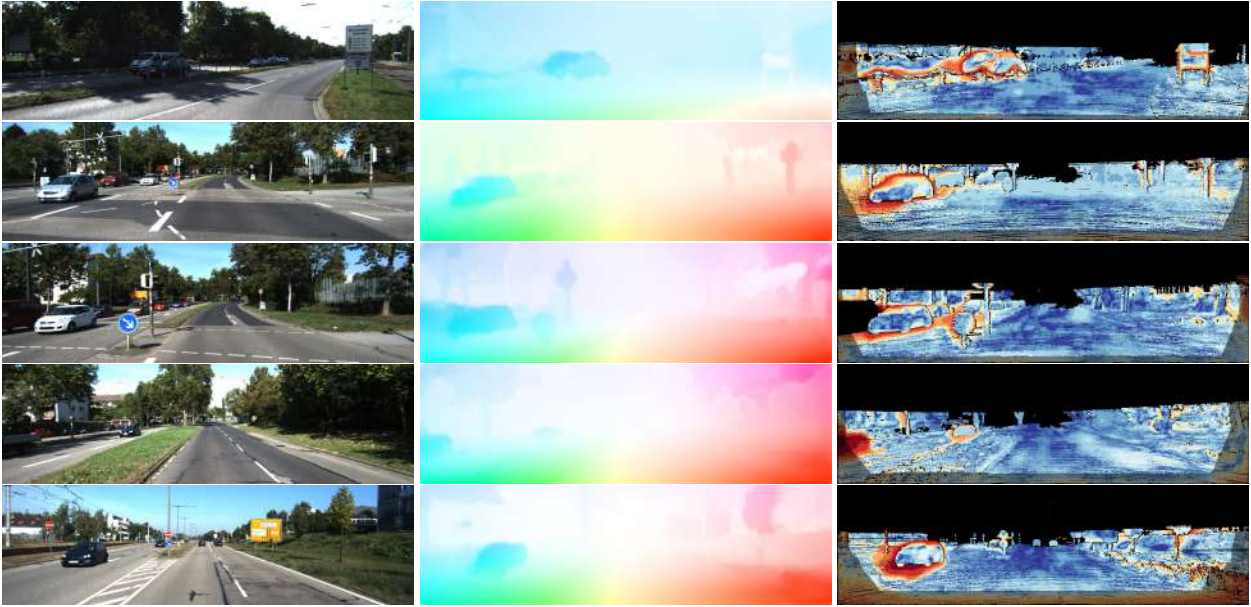

Figure 6. Visualization of the flow error map of our optical flow network on the KITTI 2015 testing benchmark. Larger errors are encoded in red, while blue pixels represents good optical flow estimates with respect to the ground-truth.

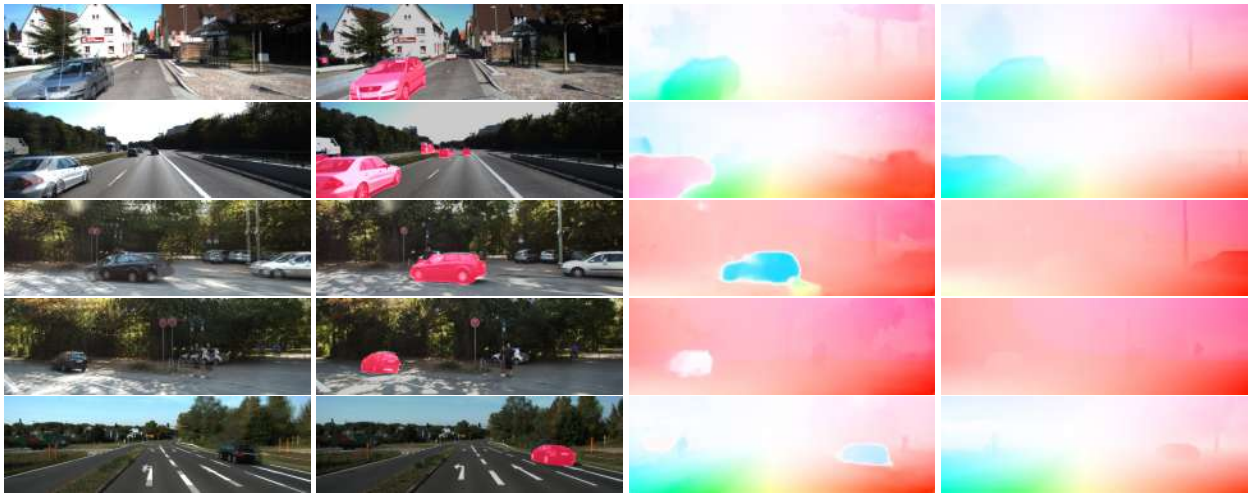

Figure 7. Qualitatives on motion segmentation. From left to right, the input images, the motion objects detected in the scene by our method (highlighted in red), the optical flow and the rigid flow.

13

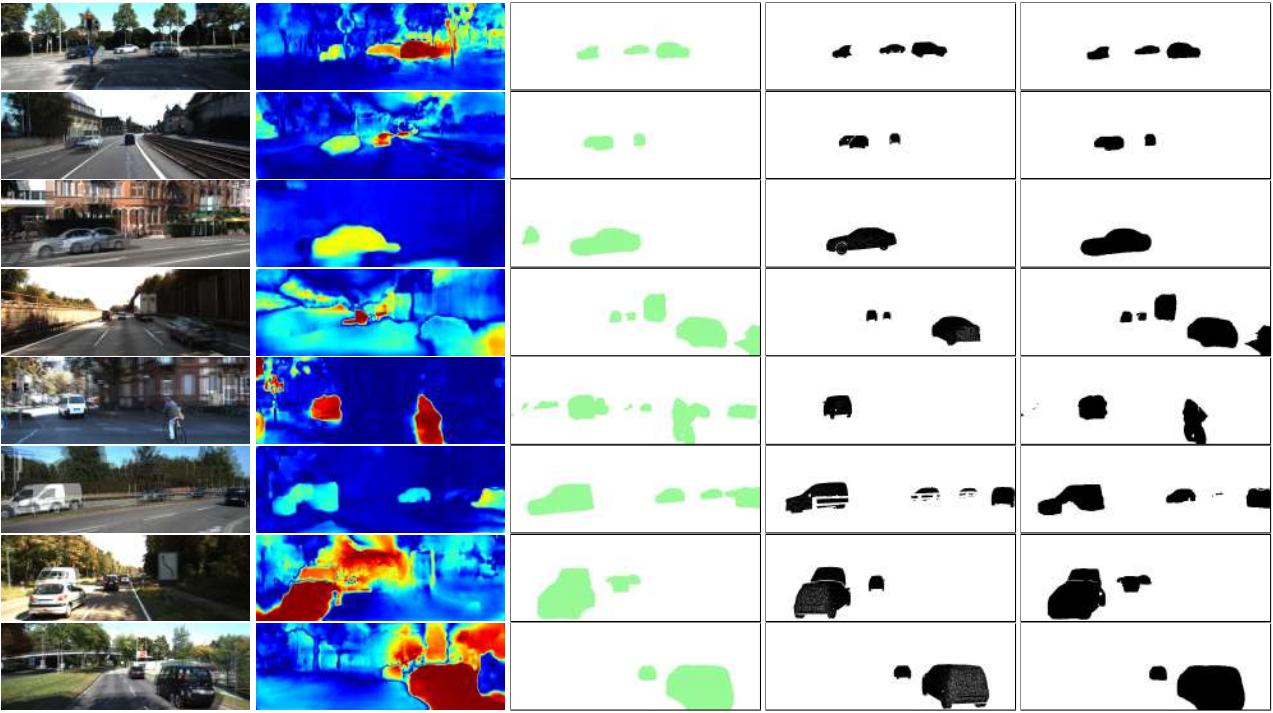

Figure 8. Motion segmentation results on the KITTI 2015 dataset. From left to right, we show the monocular sequence, the outcome of the proposed motion probability strategy (high probability of motion is encoded in red), semantic priors extracted from ours semantic predictions, ground-truth motion masks and ours estimated motion segmentation masks.

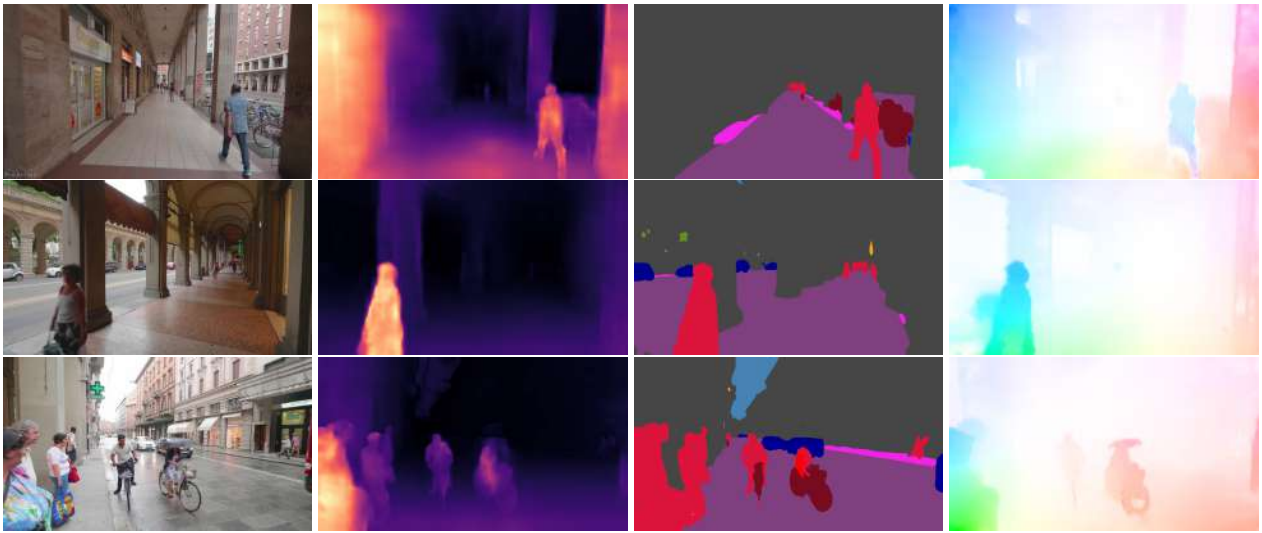

Figure 9. Qualitative results of ΩNet on a raw YouTube video. From left to right, we show the input images of a monocular sequence, the single-view depth and semantic predictions and, finally, the optical flow estimate.


}

\end{document}